%% file: main.tex
\documentclass[twoside,11pt,dvipsnames]{article}
\usepackage[preprint]{jmlr2e}

\input{math_commands.tex}

\usepackage[utf8]{inputenc}

\usepackage{hyperref}
\usepackage{url}
\usepackage{amssymb}
\usepackage{float}
\usepackage{graphicx}
\usepackage{subcaption}
\usepackage{dsfont}
\usepackage{appendix}
\usepackage{wrapfig}
\usepackage{pgfplots}

\newcommand{\pp}[1]{\left( #1 \right)}
\newcommand{\mb}{\mathbf}

\newcommand{\Cont}{{\mathcal C}}

\newcommand{\coloredbox}[2]{\colorbox{#1}{$\displaystyle #2$}}

\newcommand{\abs}[1]{\left\lvert #1 \right\rvert}
\newcommand{\norm}[1]{\left\lVert #1 \right\rVert}

\definecolor{darkgreen}{rgb}{0.0, 0.42, 0.42}

\jmlrheading{}{}{}{}{}{}{Laurent Dinh, Jascha Sohl-Dickstein, Hugo Larochelle, and Razvan Pascanu}

\ShortHeadings{A RAD approach to deep mixture models}{Dinh, Sohl-Dickstein, Larochelle, and Pascanu}
\firstpageno{1}
\begin{document}

\title{A RAD approach to deep mixture models}

\author{Laurent Dinh \email laurentdinh@google.com\\
\addr Google Brain\\
\AND
Jascha Sohl-Dickstein \email jaschasd@google.com\\
\addr Google Brain\\
\AND
Hugo Larochelle \email hugolarochelle@google.com\\
\addr Google Brain\\
\AND
Razvan Pascanu \email razp@google.com\\
\addr DeepMind\\
}

\editor{}

\maketitle

\begin{abstract}

Flow based models such as {\rmfamily\scshape Real NVP} are an extremely
powerful approach to density estimation. However, existing flow based models
are restricted to transforming continuous densities over a continuous input space into
similarly continuous distributions over continuous latent variables. This makes
them poorly suited for modeling and representing discrete
structures in data distributions, for example class membership or discrete
symmetries.
To address this difficulty,
we present a normalizing flow architecture which relies on domain partitioning using locally invertible functions, and
possesses both real and discrete valued latent variables. 
This Real and Discrete ({\rmfamily\scshape Rad}) approach retains the
desirable normalizing flow properties of exact sampling, exact
inference, and analytically computable probabilities, while at the same
time allowing simultaneous modeling of both continuous and discrete
structure in a data distribution.

\end{abstract}

\section{Introduction}

Latent generative models
 are one of the prevailing approaches
for building expressive and tractable generative models.
The generative process for a sample $\vx$ can be expressed as
\begin{align}
\vz &\sim p_Z (\vz) \\
\vx &= g(\vz),
\end{align}
where $\vz$ is a noise vector, and $g$ a parametric {\em generator network}
(typically a deep neural network).
This paradigm has several implementations, including
{\em variational autoencoders}~\citep{kingma2013auto,rezende2014stochastic} and
{\em generative adversarial networks}~\citep{goodfellow2014generative}.
Here, we base our work on {\em flow based models}~\citep{baird2005one,
tabak2013family, dinh2014nice, dinh2016density,
kingma2018glow, chen2018neural, grathwohl2018ffjord} approaches.

The training process and model architecture for many existing latent
generative models, and for all published flow based models, assumes a unimodal
smooth distribution over latent variables $\mb z$. Given the parametrization
of $g$ as a neural network, the mapping to $\mb x$ is a continuous function.
This imposed structure
 makes it challenging to model data distributions with discrete structure --
 for instance, multi-modal distributions, distributions with holes,
 distributions with discrete symmetries, or distributions that lie on a union
 of manifolds \citep[as may approximately be true for natural images,
 see][]{tenenbaum2000global}.
Indeed, such cases require the model to learn a generator whose input Jacobian
has highly varying or infinite magnitude to separate the initial noise source into different
clusters. Such variations imply a challenging optimization problem due to large
changes in curvature and introduces numerical instabilities into actual computation
of log-likelihood~\citep{behrmanninvertibility}.
This shortcoming can be critical as several problems of interest are
hypothesized to follow a clustering structure, i.e. the distributions is
concentrated along several disjoint connected sets~\citep{eghbal2018mixture}.

A standard way to address this issue has been to use
{\em mixture models}~\citep{yeung2017tackling, richardson2018gans, eghbal2018mixture}
or structured priors \citep{johnson2016composing}.
In order to efficiently parametrize the model,
mixture models are often formulated as a
{\em discrete latent variable models}~\citep{hinton2006reducing, courville2011spike, mnih2014neural, van2017neural},
some of which can be expressed as a {\em deep mixture model}~\citep{tang2012deep, van2014factoring, van2015locally}.
Although the resulting exponential number of mixture components with depth in
deep mixture models is an advantage in terms of expressivity, it is an impediment
to inference, evaluation, and training of such models, often requiring as a
result the use of approximate methods like {\em hard-{\rmfamily\scshape Em}} or
variational inference~\citep{neal1998view}.

In this paper we combine piecewise invertible functions with discrete
auxiliary variables, selecting which invertible function applies, to describe a
deep mixture model.
This framework enables a probabilistic model's latent space to have both real
and discrete valued units, and to capture both continuous and discrete
structure in the data distribution. It achieves this added capability while
preserving the exact inference, exact sampling, exact evaluation of
log-likelihood, and efficient training that make standard flow based models
desirable.

\section{Model definition}
We aim to learn a parametrized distribution $p_X(\vx)$
on the continuous input
domain $\sR^d$ by maximizing log-likelihood. The major obstacle to training an
expressive probabilistic model is typically efficiently evaluating
log-likelihood.
\begin{figure}[t]
\captionsetup[subfigure]{justification=centering}
  \centering
  \begin{subfigure}[t]{0.24\textwidth}
    \centering
    \includegraphics[height=.8\textwidth]{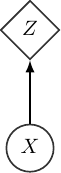}
    \caption{Inference graph for flow based model.}
  \label{fig:flow_inference}
  \end{subfigure}
  \begin{subfigure}[t]{0.24\textwidth}
    \centering
    \includegraphics[height=.8\textwidth]{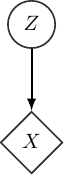}
    \caption{Sampling graph for flow based model.}
  \label{fig:flow_sampling}
  \end{subfigure}
  \begin{subfigure}[t]{0.24\textwidth}
    \centering
    \includegraphics[height=.8\textwidth]{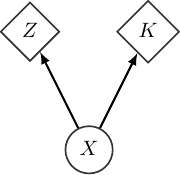}
    \caption{Inference graph for {\rmfamily\scshape Rad} model.}
  \label{fig:rad_inference}
  \end{subfigure}
  \begin{subfigure}[t]{0.24\textwidth}
    \centering
    \includegraphics[height=.8\textwidth]{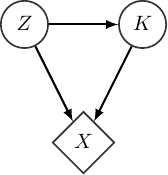}
    \caption{Sampling graph for {\rmfamily\scshape Rad} model.}
  \label{fig:rad_sampling}
  \end{subfigure}
   \caption{Stochastic computational graphs for inference and sampling for flow
   based models~(\ref{fig:flow_inference},~\ref{fig:flow_sampling}) and a 
    {\rmfamily\scshape Rad} model~(\ref{fig:rad_inference},~\ref{fig:rad_sampling}).
    Note the dependency of $K$ on $Z$ in~\ref{fig:rad_sampling}. While this is not necessary,
    we will exploit this structure as highlighted later in the main text and in
    Figure~\ref{fig:gating}.
    }
   \label{fig:scg_rad}
\end{figure}

\subsection{Partitioning}
If we consider a mixture model with a large number $\abs{K}$ of components,
where $\abs{K}$ is the number of values $K$ takes, the
likelihood takes the form
\[
p_X(\vx) = \sum_{k=1}^{\abs{K}}{p_K(k) p_{X \mid K}(\vx \mid k)}.
\]
In general, evaluating the likelihood requires computing probabilities for all $\abs{K}$
components. However, following a strategy similar to~\citet{rainforth2018it, cundy2020flexible, muller2019neural, durkan2019neural,DolatabadiEL20}, if
we partition the domain $\sR^d$ into disjoint subsets $\sA_k$ for $1 \leq k \leq \abs{K}$ such that
$\forall i \neq j~~\sA_i \cap \sA_j = \varnothing$ and $\bigcup\limits_{k=1}^{\abs{K}} \sA_k = \sR^d$,
constrain the support of $p_{X \mid K}(\vx \mid k)$ to $\sA_k$ (i.e.
$\forall \vx \notin \sA_k, p_{X \mid K}(\vx \mid k) = 0$),
and define a set identification function $f_K\pp{\vx}$ such that 
$\forall \vx \in \sR^d, \vx \in \sA_{f_K(\vx)}$ $\Big(\text{i.e.} f_K(x) = \sum_{k}{k \cdot \mathds{1}(x\in\sA_k)}\Big)$, we can write the likelihood as
\begin{align}
p_X(\vx)
&= p_K\big(f_K(\vx)\big) p_{X \mid K}\big(\vx \mid f_K(\vx)\big).
\end{align}
This transforms the problem of summation to a search problem $\vx \mapsto f_K(\vx)$.
This can be seen as the inferential converse of a {\em stratified sampling}
strategy~\citep{rubinstein2016simulation}. 

\subsection{Change of variable formula}
\begin{figure}[t]
\captionsetup[subfigure]{justification=centering}
  \centering
  \begin{subfigure}[t]{0.95\textwidth}
    \centering
    \includegraphics[width=.6\textwidth]{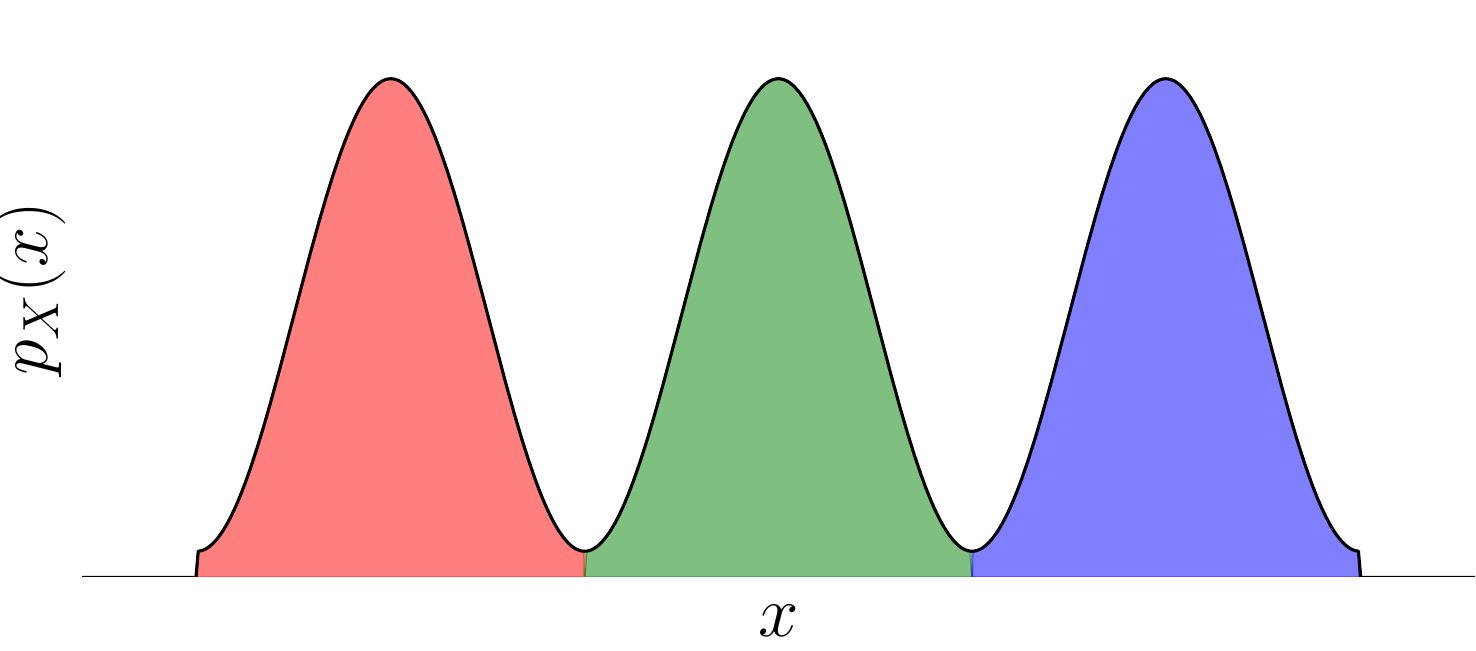}
    \caption{An example of a trimodal distribution $p_X$, sinusoidal distribution.
    The different modes are colored in red, green, and blue.}
    \label{fig:trimodal_dist}
  \end{subfigure}
  \begin{subfigure}[t]{0.58\textwidth}
    \centering
    \includegraphics[width=.6\textwidth]{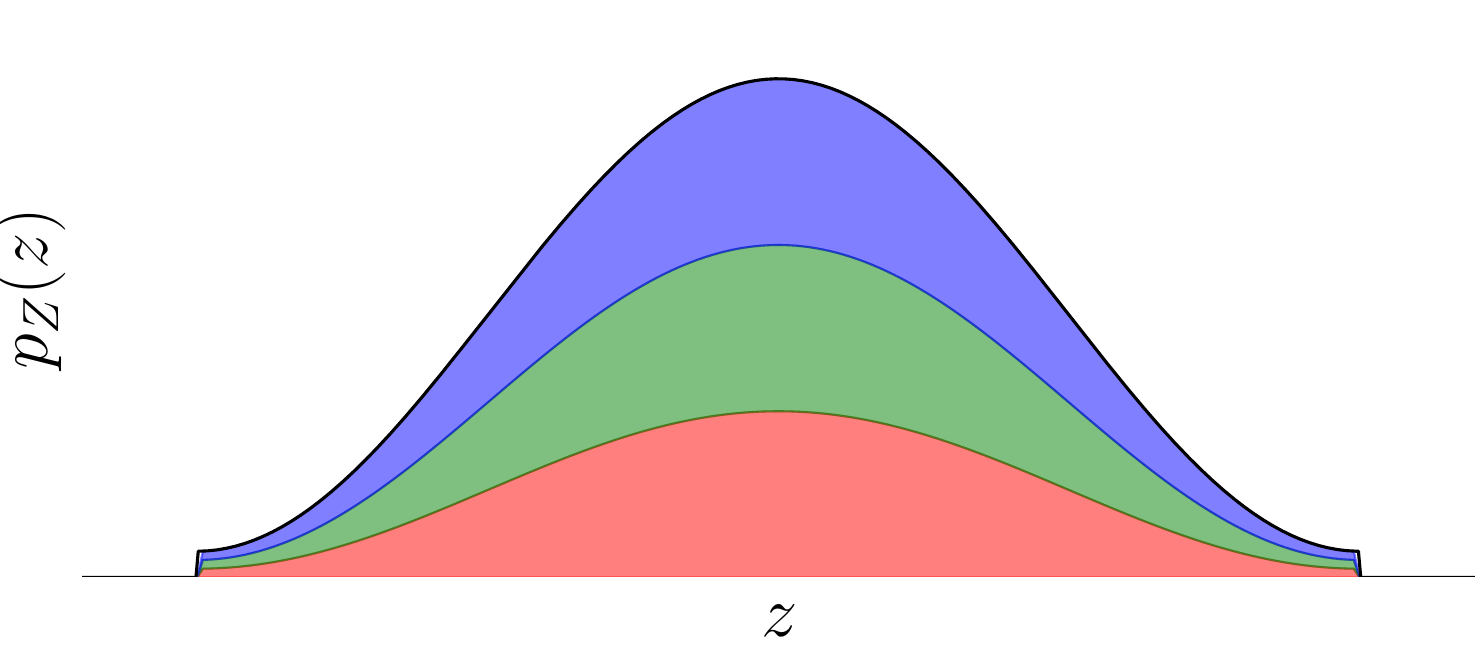}
    \caption{The resulting unimodal distribution $p_{Z}$,
    corresponding to the distribution of any of the initial modes in $p_X$.}
    \label{fig:trimodal_unimodal}
  \end{subfigure}
  \begin{subfigure}[t]{0.4\textwidth}
    \centering
    \includegraphics[width=.45\textwidth]{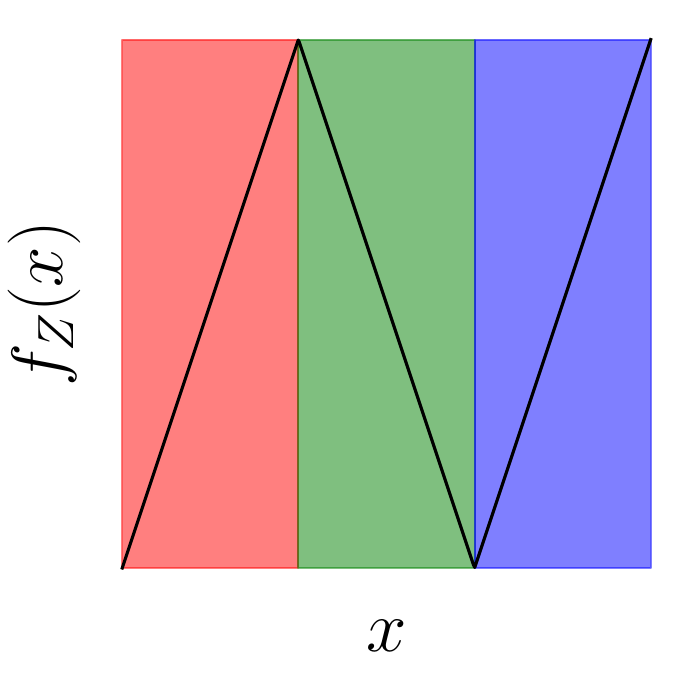}
    \caption{An example $f_{Z}(x)$ of a piecewise invertible function
    aiming at transforming $p_{Z}$ into a unimodal distribution.
    The red, green, and blue zones corresponds to the different modes in input
    space.}
    \label{fig:trimodal_map}
  \end{subfigure}
   \caption{Example of a trimodal distribution~(\ref{fig:trimodal_dist}) turned
   into a unimodal distribution~(\ref{fig:trimodal_unimodal}) using a piecewise
   invertible function~(\ref{fig:trimodal_map}). The
   distribution $p_X$ correspond to an unfolding of $p_{Z}$ as
   $p_X(x) = 
   \coloredbox{White!70!Red}{p_{Z}\left(3x + 2\right) \mathds{1}\left(x \in \sA_1\right)}
   + \coloredbox{White!70!Green}{p_{Z}(-3x) \mathds{1}\left(x \in \sA_2\right)}
   + \coloredbox{White!70!Blue}{p_{Z}\left(3x - 2\right) \mathds{1}\left(x \in \sA_3\right)}$
   where $\coloredbox{White!70!Red}{\sA_1 = \left[-1, -\frac{1}{3}\right]}$,
   $\coloredbox{White!70!Green}{\sA_2 = \left[-\frac{1}{3}, \frac{1}{3}\right]}$,
   and $\coloredbox{White!70!Blue}{\sA_3 = \left[\frac{1}{3}, 1\right]}$. Here
   $f_K(x) = \coloredbox{White!70!Red}{1\cdot\mathds{1}\left(x \in \sA_1\right)}
   + \coloredbox{White!70!Green}{2\cdot\mathds{1}\left(x \in \sA_2\right)}
   + \coloredbox{White!70!Blue}{3\cdot\mathds{1}\left(x \in \sA_3\right)}$
   }
   \label{fig:trimodal}
\end{figure}

The proposed approach will be a direct extension of flow based
models~\citep{rippel2013high,dinh2014nice,dinh2016density,kingma2018glow}.
Flow based models enable log-likelihood evaluation by relying on the
{\em change of variable formula}
\begin{align}
p_X(\vx) = p_{Z}\big(f_Z(\vx)\big) \left\lvert\frac{\partial f_Z}{\partial \vx^T}(\vx)\right\rvert,
\end{align}
with $f_Z$ a parametrized bijective function from $\sR^d$ onto
$\sR^d$ and
$\left\lvert\frac{\partial f_Z}{\partial \vx^T}\right\rvert$ the absolute value of the determinant of its
Jacobian.

As also proposed in~\citet{falorsi2019lie}, we relax the constraint that $f_Z$
be bijective, and instead have it be surjective onto $\sR^d$ and piecewise
invertible.
That is, for a partition $(\sA_k)_k$, we require $f_{Z\rvert {\sA_k}}\pp{\vx}$ to
be an invertible function, where $f_{Z\rvert {\sA_k}}\pp{\vx}$ indicates
$f_Z(\vx)$ restricted to the domain $\sA_k$.
Given a distribution $p_{Z, K}\pp{\vz, k} = p_{K\mid Z}\pp{k\mid \vz}p_{Z}\pp{\vz}$ such that
$\forall (\vz, k), \vz \notin f_Z\pp{ \sA_k} \Rightarrow p_{Z, K} = 0$,
we can define the following generative process:
\begin{align}
\vz, k &\sim p_{Z, K}(\vz, k) \\
\vx &= (f_{Z}\big\rvert_{\sA_k})^{-1}(\vz).
\end{align}
If we use the set identification function $f_K$ associated with $\sA_k$, the distribution
corresponding to this stochastic
inversion can be defined by a change of variable formula
\begin{align}
    p_X(\vx) &= \sum_{k=1}^{\abs{K}}{p_{Z, K}\big(f_Z(\vx), k\big) \left\lvert\frac{\partial f_{Z}\rvert_{\sA_k}}{\partial \vx^T}\right\rvert} \\
    &= p_{Z, K}\big(f_Z(\vx), f_K(\vx)\big) \left\lvert\frac{\partial f_{Z}}{\partial \vx^T}\right\rvert.
\end{align}
see Figure~\ref{fig:trimodal} for an example.

This contrasts with \citet{cornish2019relaxing}, which
uses instead continuous indexing for $k$ but relies as a consequence on
approximate variational inference for training their resulting model.
Because of the use of both {\em Real and Discrete} stochastic variables, we
call this class of model {\rmfamily\scshape Rad}. 
The particular parametrization we use on is depicted in Figure~\ref{fig:trimodal}.
We rely on piecewise invertible functions that allow us to
define a mixture model of repeated symmetrical patterns, following a method
of \emph{folding the input space}. In general, we use a mechanism similar
to~\citet{montufar2014number}: the non-invertibility of the surjection
enables the model to share statistical strength between the different pieces.
Note that in this instance the function $f_K$ is implicitly defined by $f_Z$,
as the discrete latent corresponds to which invertible component of the
piecewise function $\vx$ falls on.

\begin{figure}[t]
\captionsetup[subfigure]{justification=centering}
  \centering
  \begin{subfigure}[t]{0.48\textwidth}
	\centering
    \includegraphics[width=.8\textwidth]{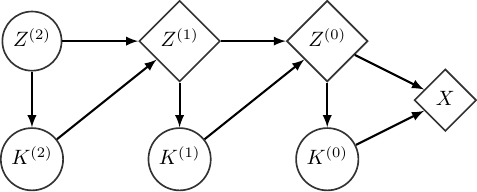}
	\caption{Sampling.}
  \end{subfigure}
  \begin{subfigure}[t]{0.48\textwidth}
	\centering
    \includegraphics[width=.8\textwidth]{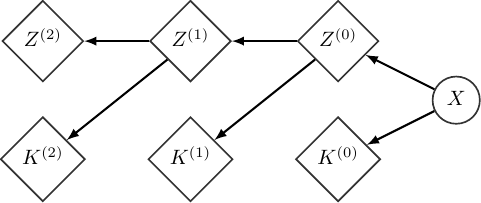}
	\caption{Inference.}
  \end{subfigure}
   \caption{Stochastic computational graph in a deep {\rmfamily\scshape Rad} mixture model of $\prod_{l=1}^{3}{\abs{K^{(l)}}}$ components.}
   \label{fig:layered_rad}
\end{figure}

\begin{figure}[t]
\captionsetup[subfigure]{justification=centering}
  \centering
  \begin{subfigure}[t]{0.95\textwidth}
	\centering
    \includegraphics[width=0.6\textwidth]{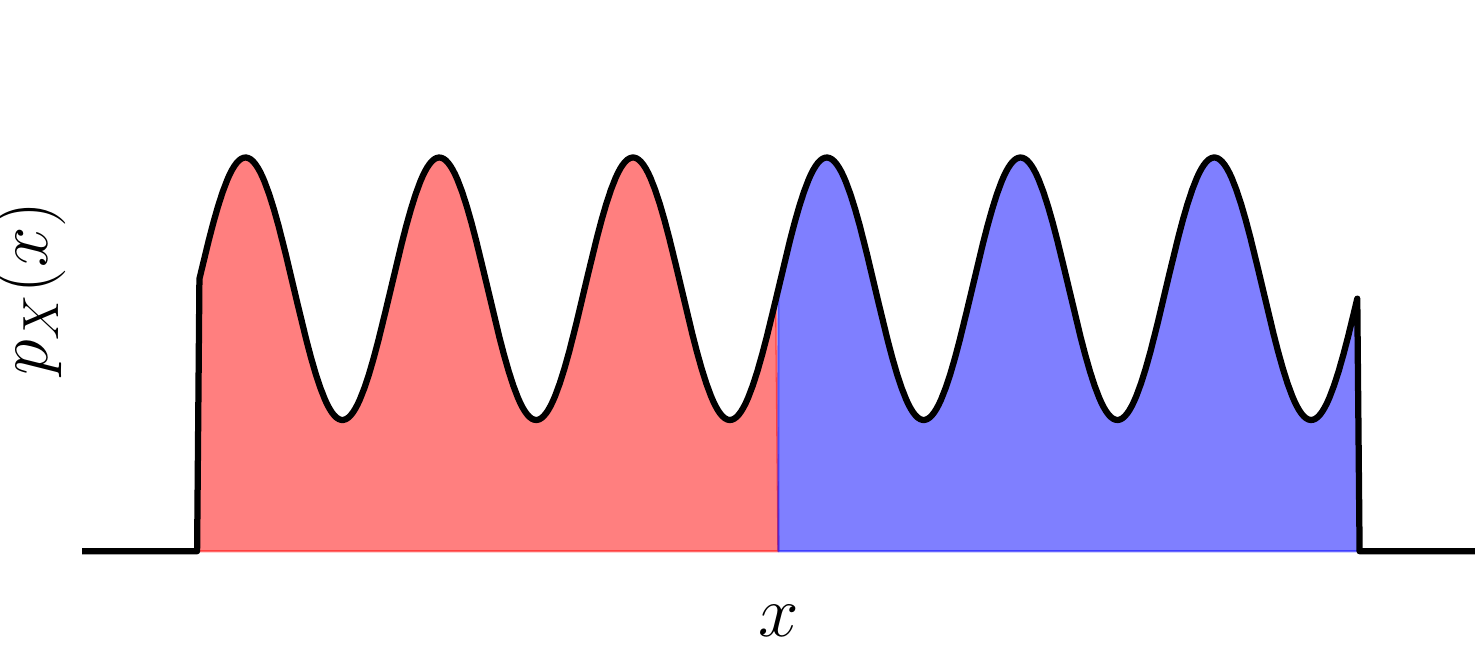}
	\caption{An example of a distribution $p_X$ (sinusoidal) with many modes.}
  \end{subfigure}
  \begin{subfigure}[t]{0.47\textwidth}
	\centering
    \includegraphics[width=0.66\textwidth]{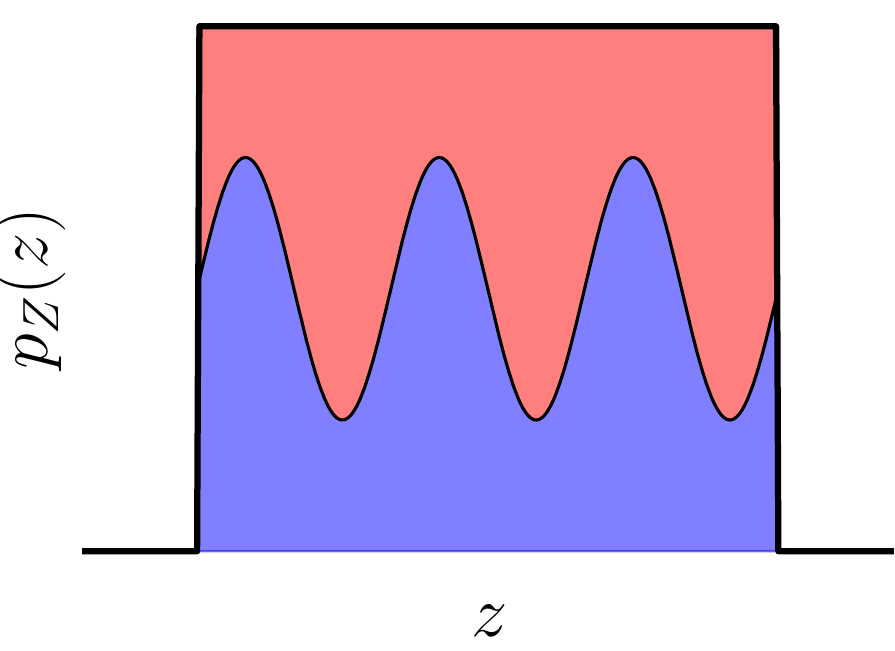}
	\caption{A simple absolute value function $z = \abs{x}$ transforms $p_X$ into $p_{Z} = \gU([0, 1])$.}
  \end{subfigure}
  \begin{subfigure}[t]{0.47\textwidth}
	\centering
    \includegraphics[width=.9\textwidth]{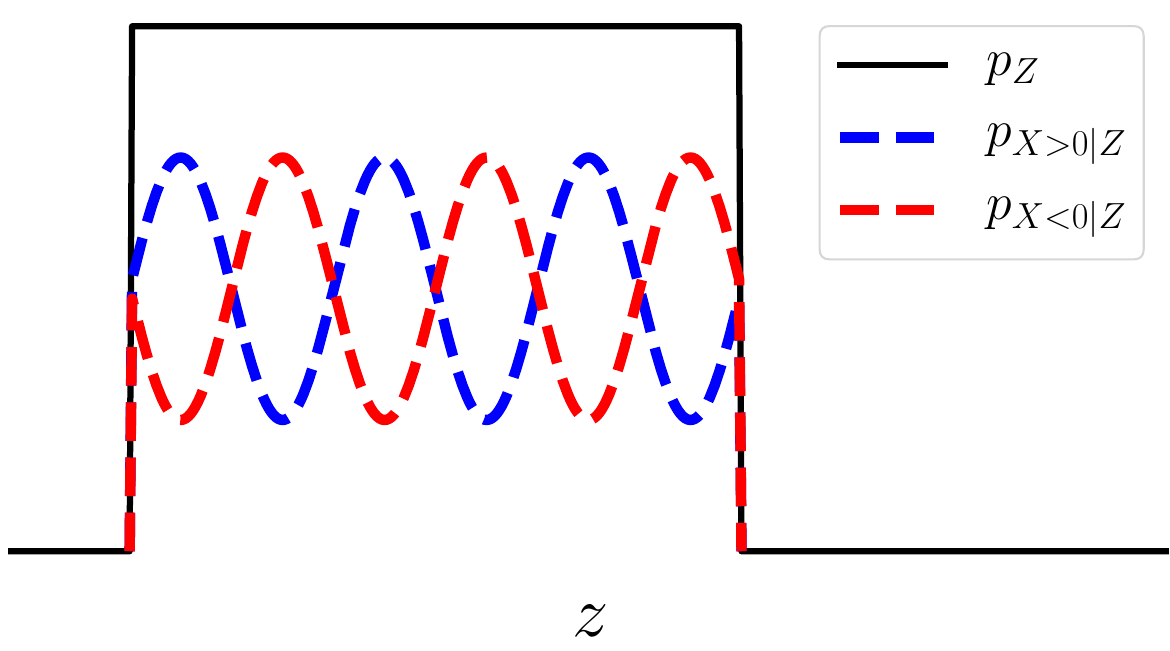}
	\caption{The {\em gating network} $p_{K \mid Z}$ allowing us to recover the
	distribution $p_X$ from $\gU([0, 1])$, here taking the form
	$\coloredbox{White!70!Red}{p_{X \leq 0 \mid Z}}$
  and $\coloredbox{White!70!Blue}{p_{X \geq 0 \mid Z}}$.}
  \end{subfigure}
   \caption{Illustration of the expressive power the gating distribution $p_{K \mid Z}$ provides.
   By capturing the structure of a sine wave in $p_{K \mid Z}$, the function
   $z, k \mapsto x$ can take on an extremely simple form, corresponding only to
   a linear function with respect to $z$. Therefore, $p_X$ does not need to be a
   duplicated version of $p_Z$.
   }
   \label{fig:gating}
\end{figure}

So far, we have defined a mixture of $\abs{K}$ components with disjoint support.
We can see in Figure~\ref{fig:trimodal} how it allows us to obtain $K$ different
modes of the distribution. However, if we factorize $p_{Z, K}$ as
$p_Z \cdot p_{K \mid Z}$, we can apply
another piecewise invertible map to $Z$ to define $p_Z$ as another multimodal
mixture model. Recursively applying this method results in a deep mixture model
(see Figure~\ref{fig:layered_rad}). 

Another advantage of such factorization is in the
{\em gating network} $p_{K \mid Z}$, as also designated in~\citep{van2015locally}.
It provides a more constrainted
but less sample wasteful approach than rejection
sampling~\citep{grover2018variational,azadi2018discriminator,bauer2018resampled} of
taking into account the untransformed sample $\vz$ before selecting the mixture
component $k$. This allows the model to exploit the distribution $p_Z$ in different
regions $\sA_k$ in more complex ways than repeating it as a pattern as illustrated in
Figure~\ref{fig:gating}.


\begin{figure}[t]
\captionsetup[subfigure]{justification=centering}
  \centering
  \begin{subfigure}[t]{0.32\textwidth}
  \centering
    \includegraphics[width=\textwidth]{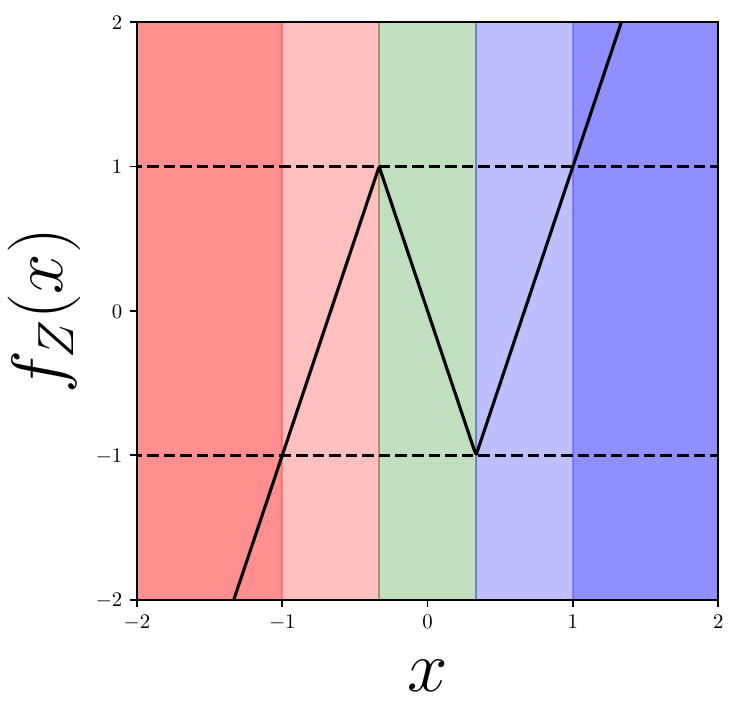}
  \caption{A simple non-invertible piecewise linear function $f_Z$ with three linear pieces
  but that cannot satisfy boundary conditions.}
  \label{fig:first_zigzag}
  \end{subfigure}
  \begin{subfigure}[t]{0.32\textwidth}
  \centering
    \includegraphics[width=\textwidth]{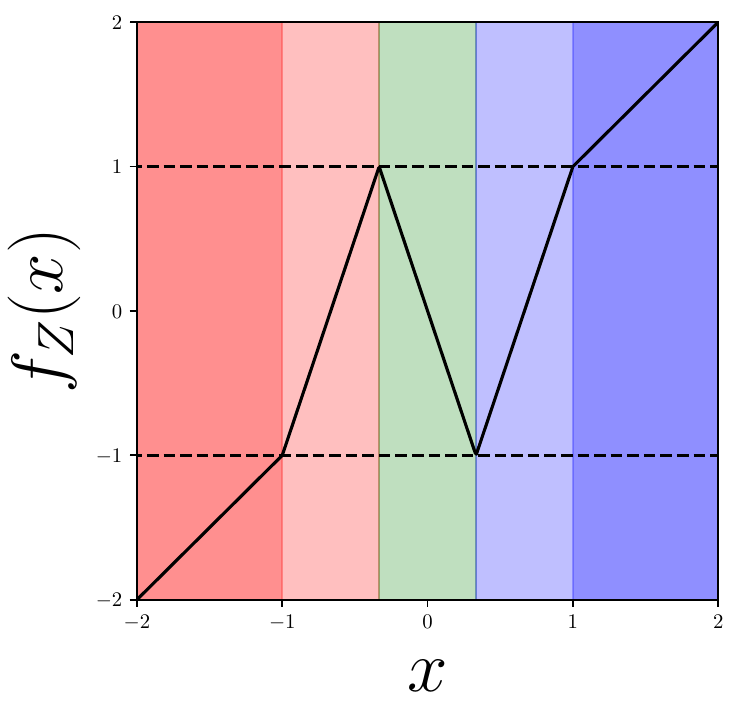}
  \caption{A simple non-invertible piecewise linear function $f_Z$ with five linear pieces
  satisfying boundary conditions.}
  \label{fig:final_zigzag}
  \end{subfigure}
  \begin{subfigure}[t]{0.32\textwidth}
  \centering
    \includegraphics[width=\textwidth]{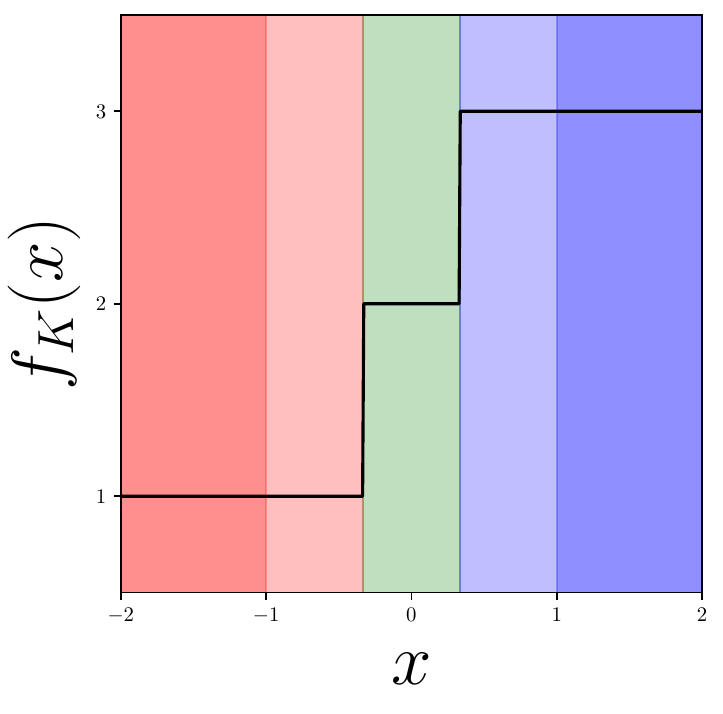}
  \caption{The $f_K$ function associated to either of these piecewise linear
  functions.}
  \label{fig:f_k}
  \end{subfigure}
   \caption{Simple piecewise linear scalar function $f_Z$
   before~(\ref{fig:first_zigzag}) and after~(\ref{fig:final_zigzag}) satisfying
   boundary conditions. The colored area correspond to the different indices
   for the mixture components, lighter color for non-invertible areas. The
   area between the dashed lines correspond to the non-invertible area in the output space.
   In~\ref{fig:f_k}, we show the $f_K$ function resulting from these nonlinearities.}
   \label{fig:zigzag}
\end{figure}

\section{Building the surjection}
\subsection{A minimal example of non-invertible surjection}
\label{sec:minimal-surjection}
As a first example of a non-invertible but piecewise invertible surjection,
we choose a scalar piecewise linear model defined as follow by
defining boundaries $a \leq b$ (see Figure
\ref{fig:first_zigzag}),
\begin{align}
    f_Z(x) = \begin{cases}
      \alpha_1 (x - a) - \alpha_2 a + c, & \text{if}\ \coloredbox{White!70!Red}{x \leq a} \\
      \alpha_3 (x - b) - \alpha_2 b + c, & \text{if}\ \coloredbox{White!70!Blue}{x \geq b} \\
      -\alpha_2 x + c, & \text{if}\ \coloredbox{White!70!Green}{x \in [a, b]}
    \end{cases}.
\end{align}
This parametrized function can represent affine functions under the
particular case $a = b$ and $\alpha_i = \alpha$.

Since an output can at most have $3$ different inverses, we have $\abs{K} = 3$.
This is one of the simplest continuous, numerically stable, and differentiable
almost everywhere surjection we can propose. Indeed, for a uniformly continous
surjective non-invertible (and therefore not stricly monotone) scalar function,
there exists a point whose preimage has at least three elements. Moreover,
each inverse of an output is simply an affine function of the output.

A non-monotonic $\Cont^1$ scalar function would have its derivative change sign
and have its log-derivative, which is ultimately a contribution to the
log-likelihood, go to $-\infty$ (according the intermediate value theorem). A
function which would admit $\pm \infty$ as a limit on a point might suffer from
numerical instability.

The image, the set of outputs, of a parametrized non-surjective function is
in general cannot be defined compactly enough to define a simple distribution
over this image, the image of a deep rectified network for example.
If the support of $p_Z$ is larger than this image, the inverse is either not
defined or has to be replaced by an improvised inverse function, resulting in a
lower bound on the actual generative model log-likelihood. For example, there
is no inverse of the absolute value on $\sR_-$. However, one could use the
identity function and $z \mapsto -z$ as makeshift inverses of the absolute
value function on $\sR_-$. This means that $x$ have now two points of origin
$z = \abs{x}$ and $z = -\abs{x}$, meaning that only using $z = \abs{x}$
results in a lower bound on the log-likelihood. On other hand, taking into
account all the points of origin of $x$ and summing their contributions into
the likelihood takes us back to the problem of evaluating many mixture
components.

Through affine functions before and after $f_Z$, we can pick without loss of
generality a normalized version of $f_Z$ where at the boundaries
$f_Z(a) = -f_Z(b) = 1$, $f_Z(1) = -f_Z(-1) = 1$, and $-1 < a < b < 1$
for further analyses. This implies that
\begin{align}
&\sum_{i = 1}^{3}{\alpha_i^{-1}} = 1 \\
&a = 2\alpha_1^{-1} - 1 \\
&b = 1 - 2\alpha_3^{-1} \\
&c = 1 + \alpha_2 (2 \alpha_1^{-1} - 1)
\end{align}
and therefore
\begin{align}
    f_Z(x) = \begin{cases}
      \alpha_1 (x + 1) - 1, & \text{if}\ \coloredbox{White!70!Red}{x \leq a} \\
      \alpha_3 (x - 1) + 1, & \text{if}\ \coloredbox{White!70!blue}{x \geq b} \\
      -\alpha_2 (x - a) + 1, & \text{if}\ \coloredbox{White!70!Green}{x \in [a, b]}
    \end{cases}.
\end{align}

\subsection{Continuity}
\label{sec:continuity}
\begin{figure}[t!]
\vspace{10pt}
\captionsetup[subfigure]{justification=centering}
  \centering
  \begin{subfigure}[t]{0.32\textwidth}
  \centering
    \includegraphics[width=\textwidth]{first_zigzag.pdf}
  \caption{A non-invertible piecewise linear function $f_Z$ with three linear pieces.}
  \label{fig:zigzag_broken}
  \end{subfigure}
  \begin{subfigure}[t]{0.32\textwidth}
  \centering
    \includegraphics[width=\textwidth]{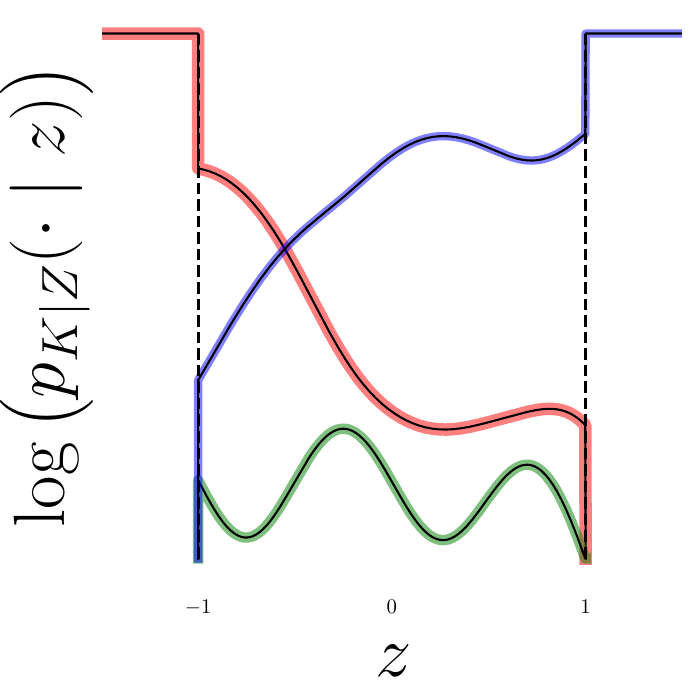}
  \caption{An example of $p_{K \mid Z}$, each curve corresponds to a
  $p_{K \mid Z}(k \mid \cdot)$ as a function of $z$.}
  \label{fig:p_k_z_broken}
  \end{subfigure}
  \begin{subfigure}[t]{0.32\textwidth}
  \centering
    \includegraphics[width=\textwidth]{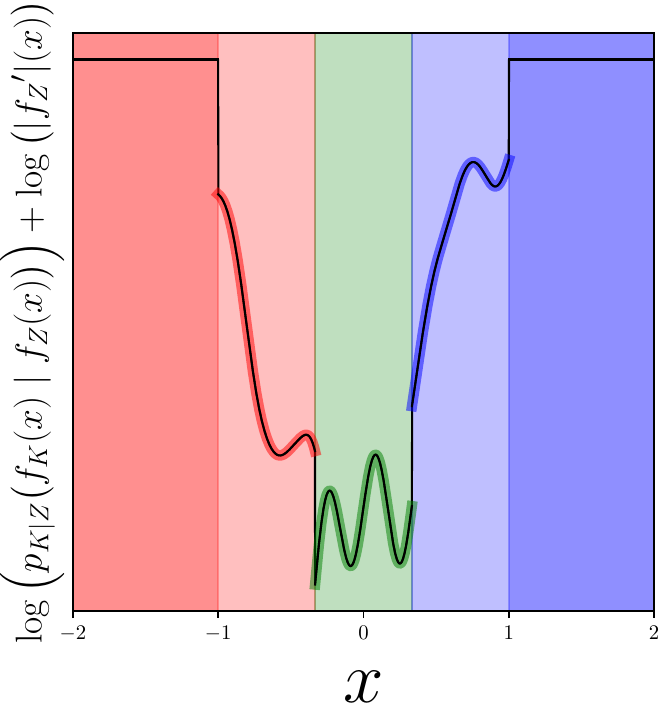}
  \caption{The resulting {\em pseudo log-Jacobian determinant}. This function, and therefore
  the resulting density $p_X$, is discontinuous.}
  \label{fig:p_k_x_broken}
  \end{subfigure}
  \begin{subfigure}[t]{0.32\textwidth}
  \centering
    \includegraphics[width=\textwidth]{final_zigzag.pdf}
  \caption{A non-invertible piecewise linear function $f_Z$ with five linear pieces.}
  \label{fig:zigzag_fixed}
  \end{subfigure}
  \begin{subfigure}[t]{0.32\textwidth}
  \centering
    \includegraphics[width=\textwidth]{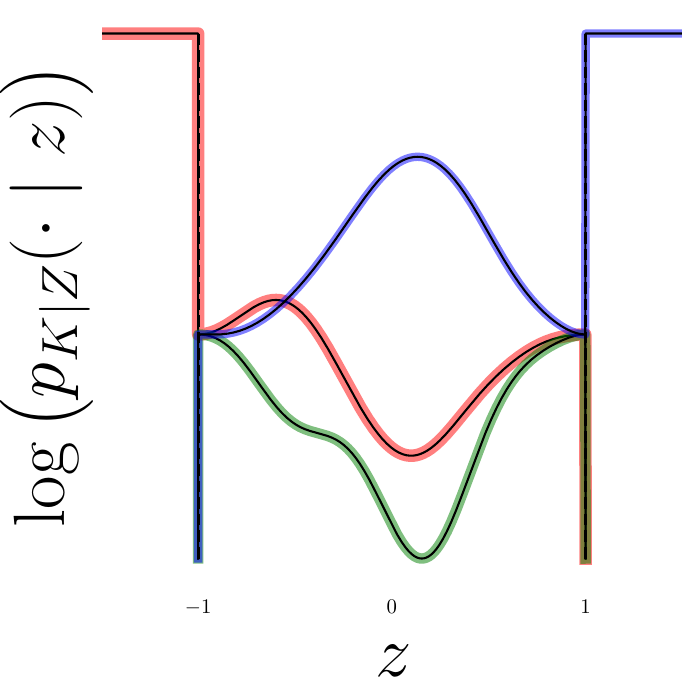}
  \caption{An example of $p_{K \mid Z}$, each curve corresponds to a
  $p_{K \mid Z}(k \mid \cdot)$ as a function of $z$. The function has been adapted as to
  respect boundary conditions.}
  \label{fig:p_k_z_fixed}
  \end{subfigure}
  \begin{subfigure}[t]{0.32\textwidth}
  \centering
    \includegraphics[width=\textwidth]{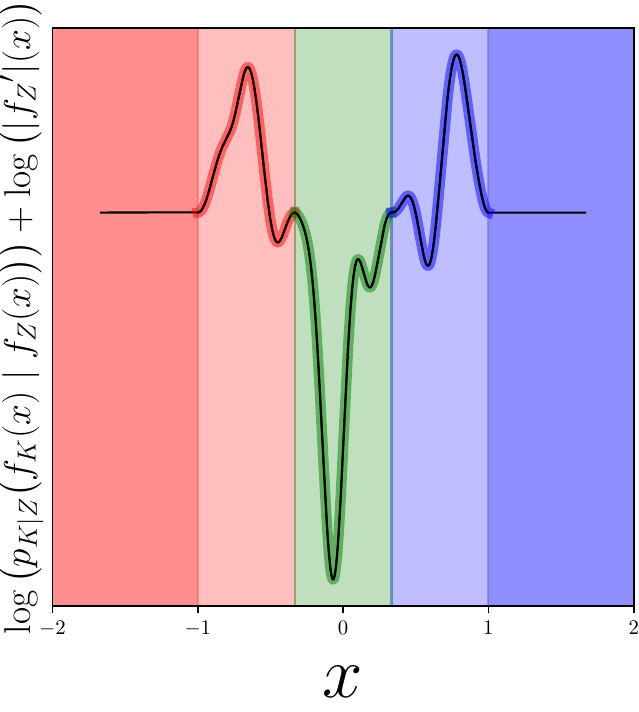}
  \caption{The resulting {\em pseudo log-Jacobian determinant}. This function, and therefore
  the resulting density $p_X$, is continuous now that the boundary conditions have been
  respected.}
  \label{fig:p_k_x_fixed}
  \end{subfigure}
   \caption{Illustration of the importance of boundary conditions to impose continuity on
   the loss function. On top, the non-invertible function $f_Z$ has three linear
   pieces~(\ref{fig:zigzag_broken}) and cannot satisfy boundary conditions. Moreover,
   $p_{K \mid Z}$~(\ref{fig:p_k_z_broken}) does not satisfy the boundary conditions
   with $f_Z$. This results in a discontinuous contribution to the
   log-likelihood~(\ref{fig:p_k_x_broken}). On the contrary, a non-invertible function
   with five pieces~(\ref{fig:zigzag_fixed}) can satisfy the boundary conditions with
   an adjusted $p_{K \mid Z}$~(\ref{fig:p_k_z_fixed}) to obtain a continuous contribution
   to the log-likelihood~(\ref{fig:p_k_x_fixed})}
   \label{fig:continuity}
\vspace{10pt}
\end{figure}

The standard approach in learning a deep probabilistic model has been
stochastic gradient descent on the negative log-likelihood. Although the model
enables the computation of a gradient almost everywhere, the log-likelihood is
unfortunately discontinuous. Let's decompose the log-likelihood
\begin{align}
\log\big(p_X(\vx)\big) = \log\left(p_Z\big(f_Z(\vx)\big)\right)
+ \log\left(p_{K \mid Z}\big(f_K(\vx) \mid f_Z(\vx)\big)\right)
+ \log\left(\left\lvert\frac{\partial f_Z}{\partial \vx^T}\right\rvert(\vx)\right).
\end{align}
There are two sources of discontinuity in this expression with respect to
the parameters and the input $\vx$ (which can be the output of another
parametrized function): $f_K$ is a function
with discrete values (therefore discontinuous, see Figure~\ref{fig:f_k} and
\ref{fig:continuity}) and
$\frac{\partial f_Z}{\partial \vx^T}$ is discontinuous because of the transitions
between the subsets $\sA_k$, leading to the expression of interest
\[
\log\left(p_{K \mid Z}\big(f_K(\vx) \mid f_Z(\vx)\big)\right)
+ \log\left(\left\lvert\frac{\partial f_Z}{\partial \vx^T}\right\rvert(\vx)\right),
\]
which takes on a role similar to that of the log-Jacobian determinant, a
{\em pseudo log-Jacobian determinant}. In the scalar case
$\left\lvert\frac{\partial f_Z}{\partial \vx^T}\right\rvert(\vx)$
is replaced by $\left\lvert f'_Z(x)\right\rvert(\vx)$.

We can attempt at parametrizing the surjection such that the pseudo log-Jacobian
determinant becomes continuous with respect to $\vx$ by expressing the
boundary condition at $x = b$
\begin{align}
&\coloredbox{White!80!Green}{\lim_{x \uparrow b} \log\left(p_{K \mid Z}\big(f_K(x) \mid f_Z(x)\big)\right)
+ \log\left(\left\lvert{f_Z}'\right\rvert(x)\right)} \\
= &\coloredbox{White!80!blue}{\lim_{x \downarrow b} \log\left(p_{K \mid Z}\big(f_K(x) \mid f_Z(x)\big)\right)
+ \log\left(\left\lvert{f_Z}'\right\rvert(x)\right)}.
\end{align}
where $\lim_{x \uparrow b}$ and $\lim_{x \downarrow b}$ are
the one-sided limits at $b$ from below and above. By remembering that
$\forall x > b, \left\lvert{f_Z}'\right\rvert(x) = \alpha_3$ and 
$f_K(x) = 3$, and
$\forall x \in ]a, b[, \left\lvert{f_Z}'\right\rvert(x) = \alpha_2$
and $f_K(x) = 2$, we obtain
\begin{align}
\Rightarrow \log&\left(p_{K \mid Z}\big(2 \mid f_Z(b)\big)\right)
- \log\left(p_{K \mid Z}\big(3 \mid f_Z(b)\big)\right)
= \log\left(\alpha_3\right)
- \log\left(\alpha_2\right),
\end{align}
Similarly, at $x= a$
\begin{align}
\log\left(p_{K \mid Z}\big(2 \mid f_Z(a)\big)\right)
- \log\left(p_{K \mid Z}\big(1 \mid f_Z(a)\big)\right)
= \log\left(\alpha_1\right)
- \log\left(\alpha_2\right).
\end{align}

Another type of boundary condition can be found at between the non-invertible
area and the invertible area, e.g. $z = 1$. When $z > 1$,
$p_{K \mid Z}\big(3 \mid z\big) = 1$, therefore the boundary condition is
\begin{align}
\lim_{x \uparrow 1} \log\left(p_{K \mid Z}\big(f_K(x) \mid f_Z(x)\big)\right)
+ &\log\left(\left\lvert{f_Z}'\right\rvert(x)\right) \\
= \lim_{x \downarrow 1} \log\left(p_{K \mid Z}\big(f_K(x) \mid f_Z(x)\big)\right)
+ &\log\left(\left\lvert{f_Z}'\right\rvert(x\right) \\
\Rightarrow \lim_{x \uparrow 1} \log\left(p_{K \mid Z}\big(3 \mid f_Z(x)\big)\right)
= \lim_{x \downarrow 1}\log\left(\left\lvert{f_Z}'\right\rvert(x)\right)&
- \lim_{x \uparrow 1}\log\left(\left\lvert{f_Z}'\right\rvert(x)\right).
\end{align}
This condition cannot be satisfied if
$\lim_{x \downarrow 1}\left\lvert{f_Z}'\right\rvert(x) = \lim_{x \uparrow 1}\left\lvert{f_Z}'\right\rvert(x)$.
Therefore, we can add more linear pieces at the end and start to redefine our
surjection\footnote{In a more general setting, one could create adjustable
gaps in derivatives using leaky rectifier functions~\citep{maas2013rectifier}
or modifications of spline
functions~\citep{muller2019neural,durkan2019neural,DolatabadiEL20}
that deliberately create discontinuities in the derivative.}
(see Figure \ref{fig:final_zigzag}):
\begin{align}
    f_Z(x) = \begin{cases}
      \alpha_- (x + 1) - 1,  & \text{if}\ \coloredbox{White!60!Red}{x \leq -1} \\
      \alpha_+ (x - 1) + 1,  & \text{if}\ \coloredbox{White!60!blue}{x \geq 1} \\
      \alpha_1 (x + 1) - 1, & \text{if}\ \coloredbox{White!80!Red}{x \in [-1, a]} \\
      \alpha_3 (x - 1) + 1, & \text{if}\ \coloredbox{White!80!blue}{x \in [b, 1]} \\
      -\alpha_2 (x - a) + 1, & \text{if}\ \coloredbox{White!80!Green}{x \in [a, b]}
    \end{cases}.
\end{align}
With this redefinition, we obtain the satisfiable constraint 
\begin{align}
&\log\left(p_{K \mid Z}\big(3 \mid f_Z(1)\big)\right)
= \log\left(\alpha_+\right)
- \log\left(\alpha_3\right).
\end{align}
Similarly, at $z = -1$ we have 
\begin{align}
&\log\left(p_{K \mid Z}\big(1 \mid f_Z(-1)\big)\right)
= \log\left(\alpha_-\right)
- \log\left(\alpha_1\right).
\end{align}

Using these boundary contraints with the constraint $\sum_{i = 1}^{3}{\alpha_i^{-1}} = 1$
mentioned earlier, the solution to these constraints
$\alpha_-, \alpha_1, \alpha_2, \alpha_3$ and $\alpha_+$
are uniquely determined from $p_{K \mid Z}\big(\cdot \mid f_Z(1)\big)$
and $p_{K \mid Z}\big(\cdot \mid f_Z(-1)\big)$ (and vice-versa):
\begin{align}
\alpha_2 &= 1 + \frac{p_{K \mid Z}\big(1 \mid f_Z(a)\big)}{p_{K \mid Z}\big(2 \mid f_Z(a)\big)}
+ \frac{p_{K \mid Z}\big(3 \mid f_Z(b)\big)}{p_{K \mid Z}\big(1 \mid f_Z(b)\big)} \\
\alpha_1 &= \frac{p_{K \mid Z}\big(2 \mid f_Z(a)\big)}{p_{K \mid Z}\big(1 \mid f_Z(a)\big)}
\left(1 + \frac{p_{K \mid Z}\big(1 \mid f_Z(a)\big)}{p_{K \mid Z}\big(2 \mid f_Z(a)\big)} + \frac{p_{K \mid Z}\big(3 \mid f_Z(b)\big)}{p_{K \mid Z}\big(2 \mid f_Z(b)\big)}\right) \\
\alpha_3 &= \frac{p_{K \mid Z}\big(2 \mid f_Z(b)\big)}{p_{K \mid Z}\big(3 \mid f_Z(b)\big)}
\left(1 + \frac{p_{K \mid Z}\big(1 \mid f_Z(a)\big)}{p_{K \mid Z}\big(2 \mid f_Z(a)\big)} + \frac{p_{K \mid Z}\big(3 \mid f_Z(b)\big)}{p_{K \mid Z}\big(2 \mid f_Z(b)\big)}\right) \\
\alpha_- &= \alpha_1 \cdot p_{K \mid Z}\big(1 \mid f_Z(b)\big) \\
\alpha_+ &= \alpha_3 \cdot p_{K \mid Z}\big(3 \mid f_Z(a)\big).
\end{align}
Therefore, when using this nonlinearity in the context of a coupling layer (see Figure~\ref{fig:coupling}),
the only parameters of interest for a normalized $f_Z$
(such that $f_Z(a) = f_Z(1) = 1 = -f_Z(b) = -f_Z(-1)$)
are $p_{K \mid Z}\big(\cdot \mid f_Z(1)\big)$
and $p_{K \mid Z}\big(\cdot \mid f_Z(-1)\big)$, excluding the affine functions
one would like to put before and after the normalized non-invertible function.

\begin{figure}[t]
\captionsetup[subfigure]{justification=centering}
  \centering
  \includegraphics[width=.6\textwidth]{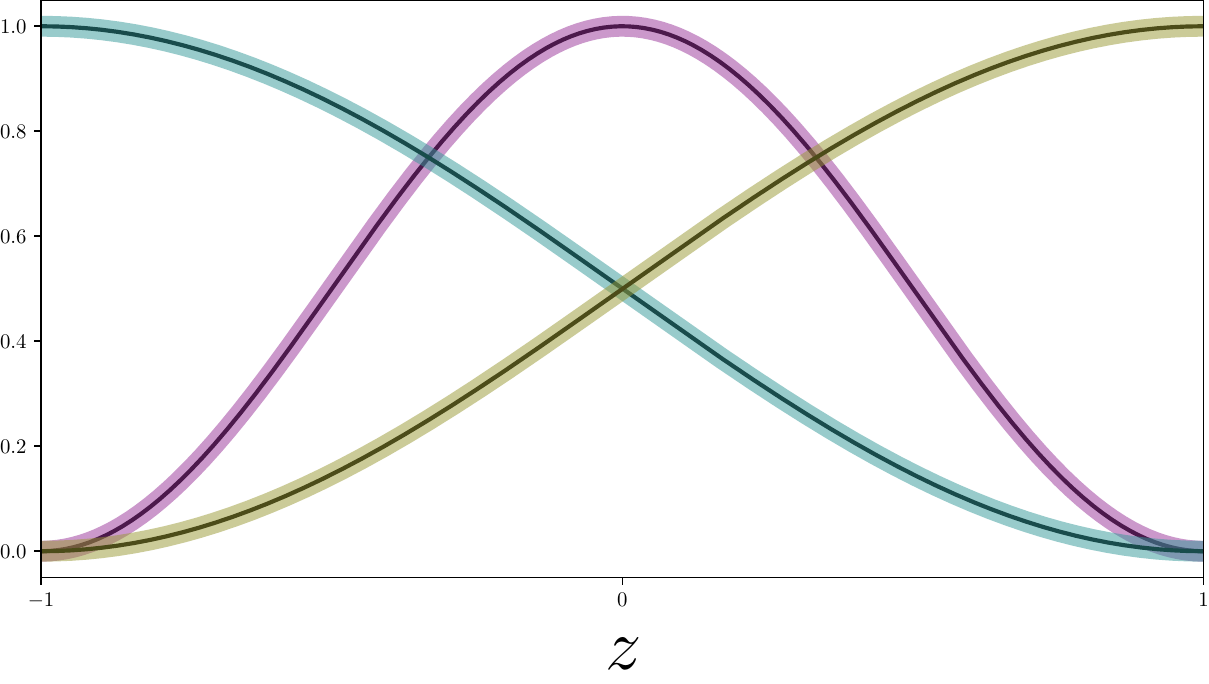}
   \caption{An example of the weightings one can put on the different
   additive components of $p_{K \mid Z}\big(\cdot \mid z\big)$. Using
   trigonometric functions, one can create weights for the 
   {\colorbox{White!70!Purple}{middle part}} (which can be arbitrary defined),
   {\colorbox{White!70!Aquamarine}{left part}} (corresponding to $p_{K \mid Z} \cdot \big(\cdot \mid f_Z(-1)\big)$),
   and {\colorbox{White!70!Dandelion}{right part}} (corresponding to $p_{K \mid Z} \cdot \big(\cdot \mid f_Z(1)\big)$).
   }
   \label{fig:weight_p_k_z}
\end{figure}

The values of $p_{K \mid Z}\big(\cdot \mid f_Z(1)\big)$ and
$p_{K \mid Z}\big(\cdot \mid f_Z(-1)\big)$ can be controlled by imposing the
following form on $p_{K \mid Z}$ as a function of its desired values at the
boundaries (see Figure~\ref{fig:weight_p_k_z})
\begin{align}
p_{K \mid Z}\big(\cdot \mid z\big)\Big\rvert_{z \in [-1, 1]}
&= {\coloredbox{White!70!Purple}{\frac{1}{2} \big(1 + \cos(z\pi)\big)}}\cdot s(z) \\
&+ {\coloredbox{White!70!Dandelion}{\frac{1}{2} \left(1 + \sin\left(\frac{z\pi}{2}\right)\right)}}\cdot p_{K \mid Z} \big(\cdot \mid f_Z(1)\big) \\
&+ {\coloredbox{White!70!Aquamarine}{\frac{1}{2} \left(1 - \sin\left(\frac{z\pi}{2}\right)\right)}}\cdot p_{K \mid Z} \big(\cdot \mid f_Z(-1)\big),
\end{align}
up to an additive constant (for normalization), where $s: \R \mapsto \R^3$
is an arbitrary parametrized function.
This reparametrization retains most of the flexibility of $p_{K \mid Z}$ while
allowing us to know its exact value at $z=-1$ and $z=1$ without evaluation, which
becomes critical in a high dimensional setting
(to avoid computating the $2d$ evaluations of
$p_{K_i \mid Z}(\cdot \mid z_i = \pm 1, z_{-i})$).
In higher dimension, $f_Z$ and $f_K$ can be applied element-wise while we can
choose $s: ~ \R^d ~ \mapsto ~ \R^{3d}$, meaning that the discrete variables $K$ are
independent conditioned on {\em all} dimensions of $\vz$. This is the approach
we use later on. Due to the discrete nature
of $f_K$ any other dependence on $K$, including auto-regressivity in $p_{K \mid Z}$
with respect to previous $K$ (but not on previous $X$),
would result in discontinuities in the log-likelihood function. Using a
mixture of conditionally independent distributions for $K$ is another
valid approach.

Given those constraints, the model can then be reliably learned through gradient
descent methods. Note that the resulting tractability of the model results from
the fact that the discrete variables $k$ is only used during inference
with the distribution $p_{K \mid Z}$, unlike discrete variational autoencoders
approaches~\citep{mnih2014neural, van2017neural} where it is fed to a deep
neural network. Similar to~\citet{rolfe2016discrete}, the learning of discrete
variables is achieved by relying on the the continuous component of the model,
and, as opposed to
other approaches~\citep{bengio2013estimating,raiko2014techniques,jang2016categorical,
maddison2016concrete,grathwohl2017backpropagation,tucker2017rebar,tran2019discrete,
hoogeboom2019integer,berg2020idf++},
this gradient signal extracted is exact and closed form.

\subsection{Universal approximation}
\label{sec:universal}
\begin{figure}[t!]
\captionsetup[subfigure]{justification=centering}
  \centering
  \begin{subfigure}[t]{0.95\textwidth}
  \centering
  \begin{tikzpicture}
    \node (img)  {\includegraphics[width=.95\textwidth]{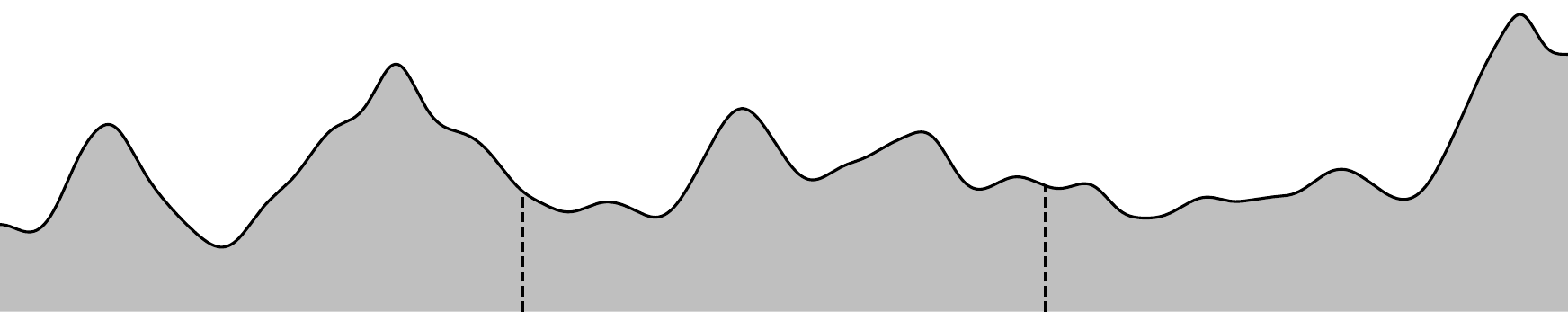}};
    \node[below of=img, node distance=1.8cm, yshift=0cm] {$x$};
    \node[left of=img, node distance=7.5cm, rotate=90, anchor=center] {$p_X(x)$};
  \end{tikzpicture}
  \caption{An example of a distribution $p_X$.}
  \vspace*{12.5pt}
  \label{fig:fold_unlabelled}
  \end{subfigure}
  \begin{subfigure}[t]{0.95\textwidth}
  \centering
  \begin{tikzpicture}
    \node (img)  {\includegraphics[width=.95\textwidth]{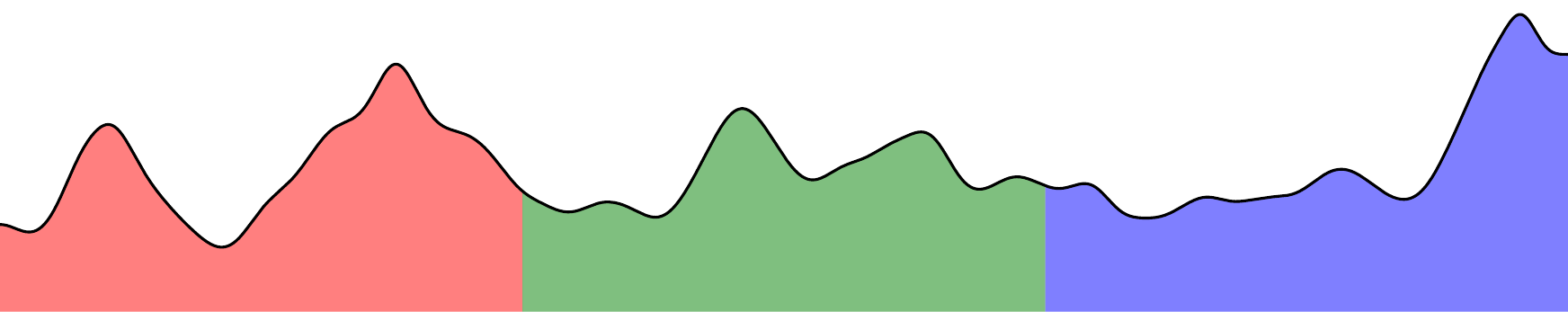}};
    \node[below of=img, node distance=1.8cm, yshift=0cm] {$x$};
    \node[left of=img, node distance=7.5cm, rotate=90, anchor=center] {$p_X(x)$};
  \end{tikzpicture}
  \caption{The same distribution $p_X$. The $x$ axis has been
  partitioned in three pieces.}
  \vspace*{12.5pt}
  \label{fig:fold_labelled}
  \end{subfigure}
  \begin{subfigure}[t]{0.95\textwidth}
  \centering
  \begin{tikzpicture}
    \node (img)  {\includegraphics[width=.95\textwidth]{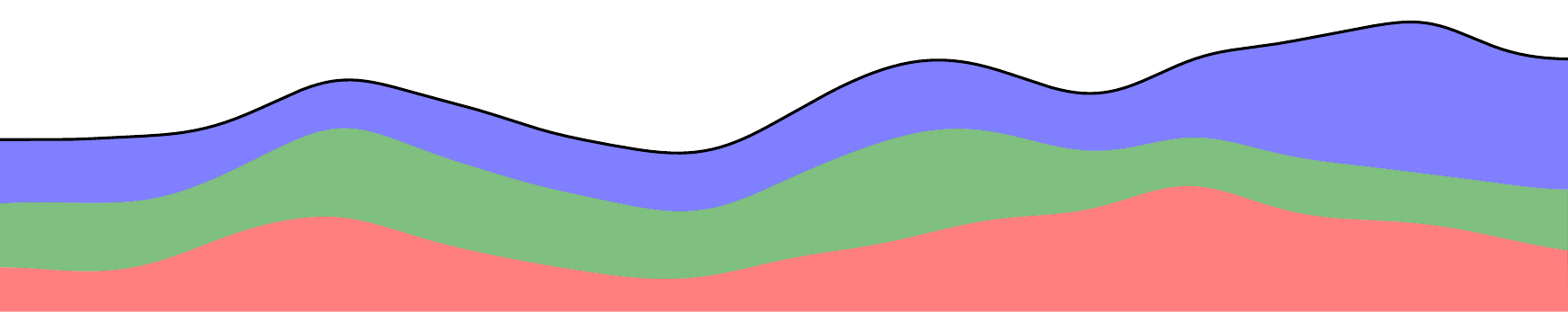}};
    \node[below of=img, node distance=1.8cm, yshift=0cm] {$z$};
    \node[left of=img, node distance=7.5cm, rotate=90, anchor=center] {$p_Z(z)$};
  \end{tikzpicture}
  \caption{The resulting smoothed out distribution $p_Z$ when applying the
  $f_Z$ described with $\alpha_1 = \alpha_2 = \alpha_3 = 3$.}
  \vspace*{12.5pt}
  \label{fig:fold_folded}
  \end{subfigure}
  \begin{subfigure}[t]{0.95\textwidth}
  \centering
  \begin{tikzpicture}
    \node (img)  {\includegraphics[width=.95\textwidth]{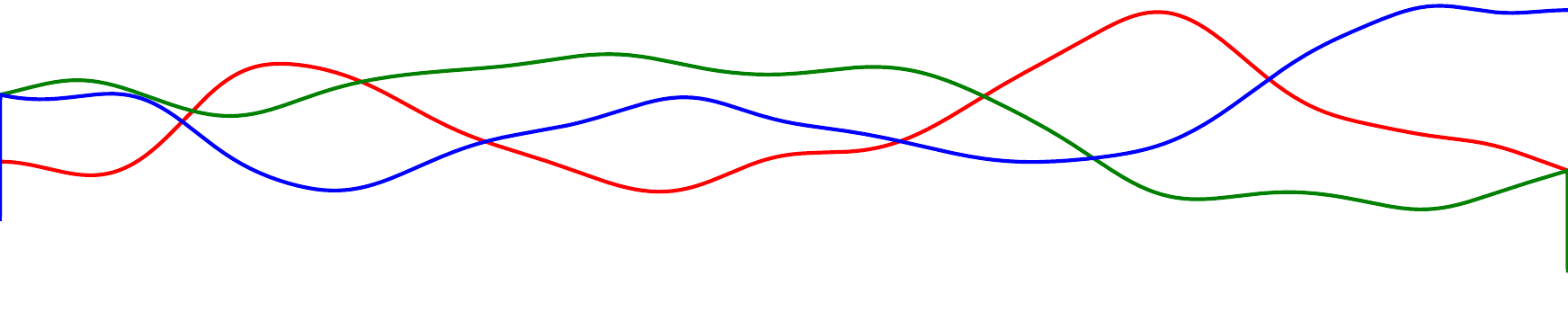}};
    \node[below of=img, node distance=1.8cm, yshift=0cm] {$z$};
    \node[left of=img, node distance=7.5cm, rotate=90, anchor=center] {$p_{K \mid Z}(\cdot \mid z)$};
  \end{tikzpicture}
  \caption{The resulting gating distribution $p_{K \mid Z}$ to recover
  the original $p_X$.}
  \vspace*{12.5pt}
  \label{fig:fold_p_k_z}
  \end{subfigure}
   \caption{Illustration of the importance of the smoothing out effect of $f_Z$.
   A more rigorous explanation of conditions for this smoothing out to be
   guaranteed is described Subsection~\ref{sec:universal}.}
   \label{fig:folding_smoothing}
\end{figure}
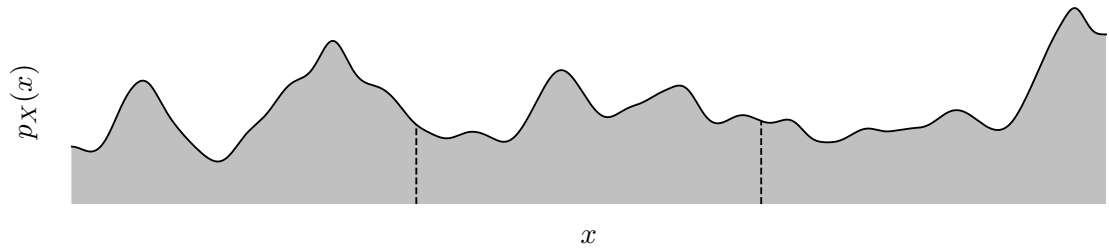
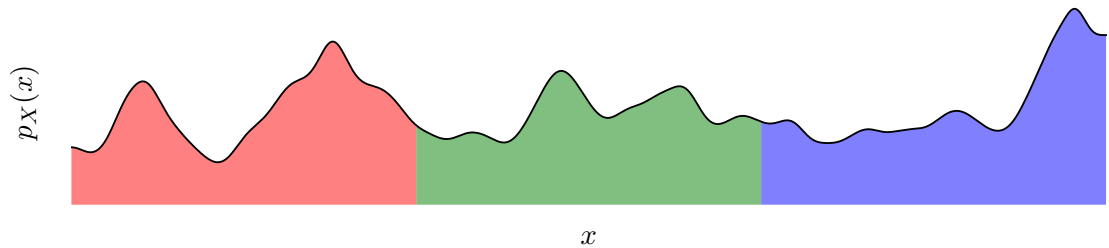
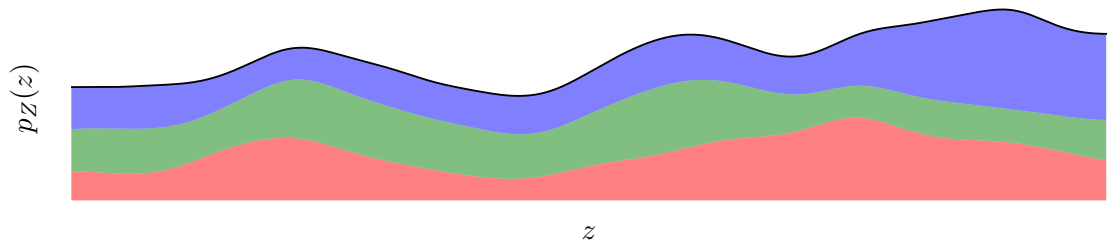
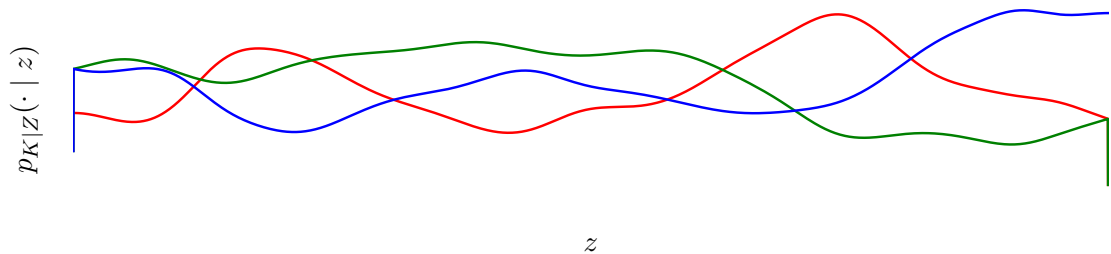

Here, we will show that this folding function $f_Z$ previously defined,
when applied repeatedly, can transform any density function so that it
becomes arbitrarily close to the uniform distribution with respect to the
uniform topology. Once, this is accomplished,
the "inversion" of $f_Z$ back to the density $p_X$ can be accomplished
by approximating the gating distribution $p_{K\mid Z}$
(a mere probabilistic classifier) arbitrarily well. Using a neural network
to converge to a given function is a well explored
topic~\citep{cybenko1989approximation, hornik1991approximation, pinkus1999approximation}.
\begin{theorem}
For any $\Cont^1([-1, 1])$ strictly positive density $p^{*}_X$ and
$\epsilon > 0$, there is a
series of surjections $(f_{Z,n})_{n \leq N}$ from $[-1, 1]$ onto $[-1, 1]$,
composed of three linear pieces, such that
\[\norm{p^{*}_{f_{Z, 1:N}(X)} - \gU\big([-1, 1]\big)}_\infty < \epsilon.\]
with $f_{Z, 1:N} = f_{Z,N} \circ \dots \circ f_{Z,1}$.

\begin{proof}
$(p^*_X)'$ is continuous on $[0, 1]$ and therefore bounded
(by the extreme value theorem). If we apply the previously defined surjection
$f_Z$ with $\alpha_1 = \alpha_2 = \alpha_3 = 3$ on a distribution $p_X$ we
obtain
\begin{align}
p_Z(z) =
\frac{1}{3}\left(\coloredbox{White!70!Red}{p_X\left(\frac{z - 2}{3}\right)}
+ \coloredbox{White!70!Green}{p_X\left(-\frac{z}{3}\right)}
+ \coloredbox{White!70!Blue}{p_X\left(\frac{z + 2}{3}\right)}\right).
\end{align}
The Lipschitz bound of $p_Z$ with respect to $z$ is a third of
$\norm{(p^*_X)'}_\infty$ see Figure~\ref{fig:folding_smoothing}.
Therefore, by repeating that process $N$ times we have
\begin{align}
&N \geq \left\lceil\log_3\left(\frac{2 \cdot \norm{(p^*_X)'}_\infty}{\epsilon}\right)\right\rceil \\
\Rightarrow &\norm{(p^{*}_{f_{Z,1:N}(X)})'}_\infty \leq \frac{1}{2}\epsilon
\end{align}
Using Rolle's theorem on the cumulative distribution function corresponding to
$p^{*}_{f_{Z,1:N}(X)}$, there is a $z \in ]-1, 1[$ such that
$p^{*}_{f_{Z,1:N}(X)}(z) = \frac{1}{2}$. Therefore
\begin{align}
\norm{p^{*}_{f_{Z,1:N}(X)} - \gU\big([-1, 1]\big)}_\infty < \epsilon.
\end{align}
\end{proof}
\end{theorem}

To recover the original distribution $p^{*}_X$ from $p^{*}_{f_{Z,1:N}(X)}$,
one can use the universal approximation property of neural
networks to parametrize gating distributions $p_{K\mid Z}$. Here, we use
uniform pieces in $f_Z$ for simplicity of proof, this can be of course a
suboptimal choice and the use of different sizes may allow us to use a
lower $N$.

\section{Experiments}
\subsection{Problems}
\begin{figure}[t]
\captionsetup[subfigure]{justification=centering}
  \centering
  \begin{subfigure}[t]{0.16\textwidth}
	\centering
    \includegraphics[width=.6\textwidth]{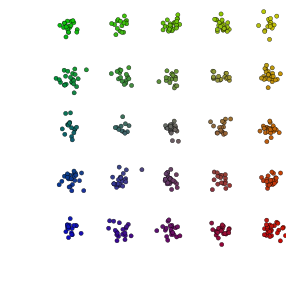}
	\caption{Grid Gaussian mixture. This problem follows the clustering
	hypothesis.}
  \end{subfigure}
  \begin{subfigure}[t]{0.16\textwidth}
	\centering
    \includegraphics[width=.6\textwidth]{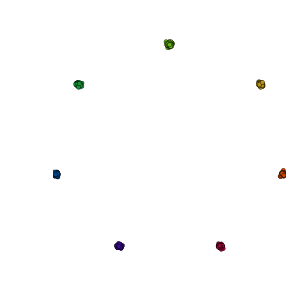}
	\caption{Ring Gaussian mixture. This problem also follows the clustering
	hypothesis but the clusters are not axis-aligned.}
  \end{subfigure}
  \begin{subfigure}[t]{0.16\textwidth}
	\centering
    \includegraphics[width=.6\textwidth]{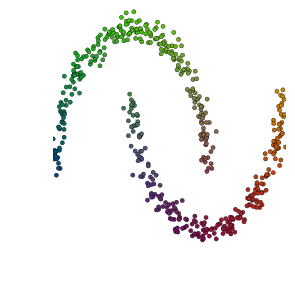}
	\caption{Two moons. This problem not only follows the clustering hypothesis
	but also the manifold hypothesis. }
  \end{subfigure}
  \begin{subfigure}[t]{0.16\textwidth}
	\centering
    \includegraphics[width=.6\textwidth]{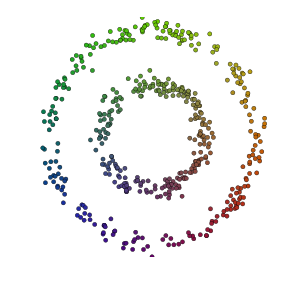}
	\caption{Two circles. This problem also follows both the clustering and
	manifold hypothesis. A continuous bijection cannot linearly separate those two
	clusters.}
  \end{subfigure}
  \begin{subfigure}[t]{0.16\textwidth}
	\centering
    \includegraphics[width=.6\textwidth]{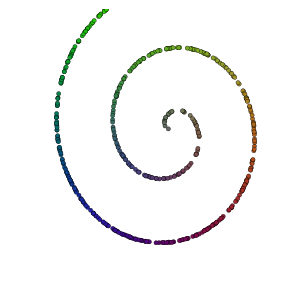}
	\caption{Spiral. This problem follows only the manifold hypothesis.}
  \end{subfigure}
  \begin{subfigure}[t]{0.16\textwidth}
	\centering
    \includegraphics[width=.6\textwidth]{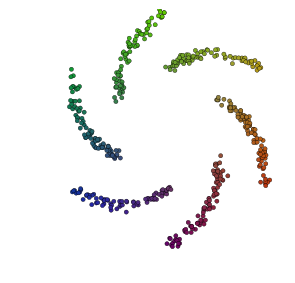}
	\caption{Many moons. This problem follows both the clustering and manifold
	hypotheses, with many clusters.}
  \end{subfigure}
   \caption{Samples drawn from the data distribution in each of several toy two dimensional problems.}
   \label{fig:toy_2d}
\end{figure}

We conduct a brief comparison on six two-dimensional toy problems with
{\rmfamily\scshape Real NVP} to demonstrate the potential gain in expressivity
{\rmfamily\scshape Rad} models can enable. Synthetic datasets of $10,000$ points
each are constructed following the {\em manifold hypothesis} and/or the
{\em clustering hypothesis}. We designate these problems as:
{\em grid Gaussian mixture},
{\em ring Gaussian mixture},
{\em two moons},
{\em two circles},
{\em spiral}, and
{\em many moons} (see Figure~\ref{fig:toy_2d}).

\begin{figure}[t]
\centering
  \begin{subfigure}[t]{0.48\textwidth}
  \centering
  \includegraphics[width=.6\textwidth]{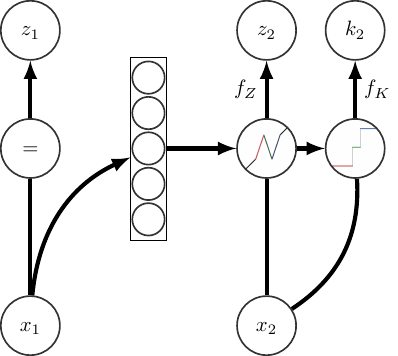}
  \caption{Forward pass.}
  \end{subfigure}
  \begin{subfigure}[t]{0.48\textwidth}
  \centering
  \includegraphics[width=.6\textwidth]{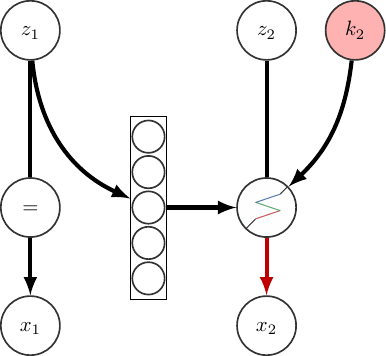}
  \caption{Inversion graph.}
  \end{subfigure}
  \caption{Computational graph of the coupling layers used in the experiments.}
  \label{fig:coupling}
\end{figure}

\subsection{Architecture}

For the {\rmfamily\scshape Rad} model implementation, we use the piecewise
linear activations defined in Subsection~\ref{sec:continuity} in a coupling layer
architecture~\citep{dinh2014nice, dinh2016density} for $f_Z$ where, instead of a
conditional linear transformation, the conditioning variable $x_1$ determines
the parameters of the piecewise linear activation on $x_2$ to obtain $z_2$ and
$k_2$, with $z_1 = x_1$ (see Figure~\ref{fig:coupling}). For the gating network $p_{K \mid Z}$, the gating logit
neural network $s\pp{z}$ take as input $z = (z_1, z_2)$. We compare with a
{\rmfamily\scshape Real NVP} model using only affine coupling layers. $p_Z$ is
a standard Gaussian distribution.

As both these models can easily approximately solve these generative modeling tasks provided
enough capacity, we study these model in a relatively low capacity regime,
where we can showcase the potential expressivity {\rmfamily\scshape Rad} may
provide. Each of these models uses six coupling layers, and each coupling
layer uses a one-hidden-layer rectified network with a $\tanh$ output
activation scaled by a scalar parameter
as described in~\citet{dinh2016density}. For {\rmfamily\scshape Rad}, the logit
network $s\pp{\cdot}$ also uses a one-hidden-layer rectified neural network, but with linear
output. In order to
fairly compare with respect to number of parameters, we provide
{\rmfamily\scshape Real NVP} seven times more hidden units per hidden layer
than {\rmfamily\scshape Rad}, which uses $8$ hidden units per hidden layer.
For each level, $p_{K \mid Z}$ and $f_Z$ are trained using stochastic gradient ascent with
{\rmfamily\scshape Adam}~\citep{kingma2014adam} on the log-likelihood with a
batch size of $500$ for $50,000$ steps.

\subsection{Results}
In each of these problems, {\rmfamily\scshape Rad} is consistently able to
obtain higher log-likelihood than {\rmfamily\scshape Real NVP}.

\begin{center}
  \begin{tabular}{ | c | c | c | }
    \hline
    ~ & {\rmfamily\scshape Rad} & {\rmfamily\scshape Real NVP} \\ \hline
    Grid Gaussian mixture & $-1.20$ & $-2.26$ \\ \hline
    Ring Gaussian mixture & $3.57$ & $1.85$ \\ \hline
    Two moons & $-1.21$ & $-1.48$ \\ \hline
    Two circles & $-1.81$ & $-2.17$ \\ \hline
    Spiral & $0.29$ & $-0.36$ \\ \hline
    Many moons & $-0.83$ & $-1.50$ \\
    \hline
  \end{tabular}
\end{center}

\subsubsection{Sampling and Gaussianization}

\begin{figure}[t]
\captionsetup[subfigure]{justification=centering}
  \centering
  \begin{subfigure}[t]{0.16\textwidth}
	\centering
    \includegraphics[width=.8\textwidth]{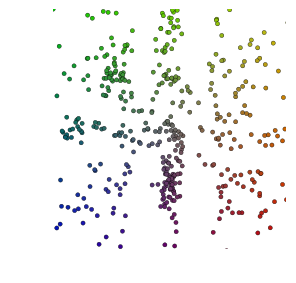}
	\caption{{\rmfamily\scshape Real NVP} on grid Gaussian mixture.}
  \end{subfigure}
  \begin{subfigure}[t]{0.16\textwidth}
	\centering
    \includegraphics[width=.8\textwidth]{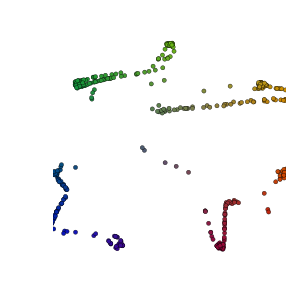}
	\caption{{\rmfamily\scshape Real NVP} on ring Gaussian mixture.}
  \end{subfigure}
  \begin{subfigure}[t]{0.16\textwidth}
	\centering
    \includegraphics[width=.8\textwidth]{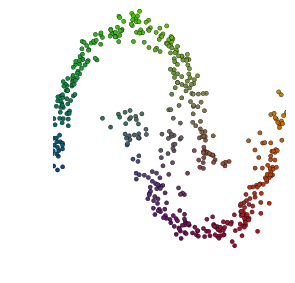}
	\caption{{\rmfamily\scshape Real NVP} on two moons.}
  \end{subfigure}
  \begin{subfigure}[t]{0.16\textwidth}
	\centering
    \includegraphics[width=.8\textwidth]{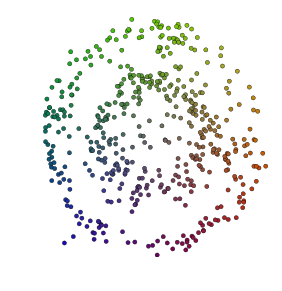}
	\caption{{\rmfamily\scshape Real NVP} on two circles.}
  \end{subfigure}
  \begin{subfigure}[t]{0.16\textwidth}
	\centering
    \includegraphics[width=.8\textwidth]{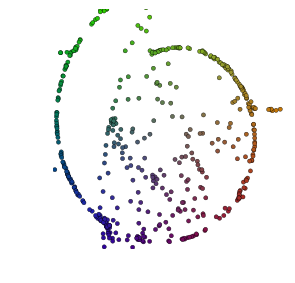}
	\caption{{\rmfamily\scshape Real NVP} on spiral.}
  \end{subfigure}
  \begin{subfigure}[t]{0.16\textwidth}
	\centering
    \includegraphics[width=.8\textwidth]{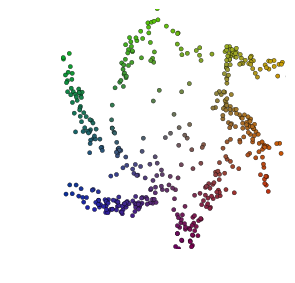}
	\caption{{\rmfamily\scshape Real NVP} on many moons.}
  \end{subfigure}

  \begin{subfigure}[t]{0.16\textwidth}
	\centering
    \includegraphics[width=.8\textwidth]{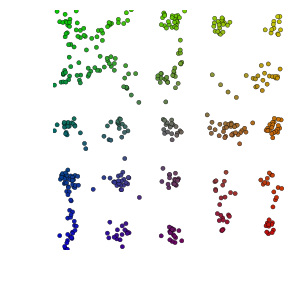}
	\caption{{\rmfamily\scshape Rad} on grid Gaussian mixture.}
  \end{subfigure}
  \begin{subfigure}[t]{0.16\textwidth}
	\centering
    \includegraphics[width=.8\textwidth]{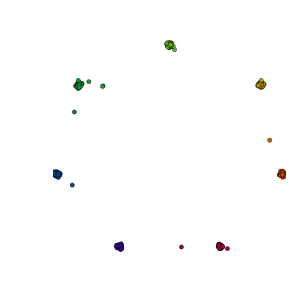}
	\caption{{\rmfamily\scshape Rad} on ring Gaussian mixture.}
  \end{subfigure}
  \begin{subfigure}[t]{0.16\textwidth}
	\centering
    \includegraphics[width=.8\textwidth]{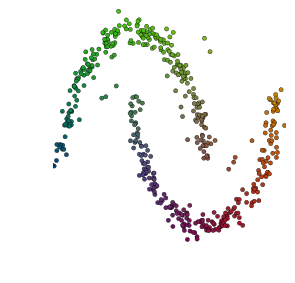}
	\caption{{\rmfamily\scshape Rad} on two moons.}
  \end{subfigure}
  \begin{subfigure}[t]{0.16\textwidth}
	\centering
    \includegraphics[width=.8\textwidth]{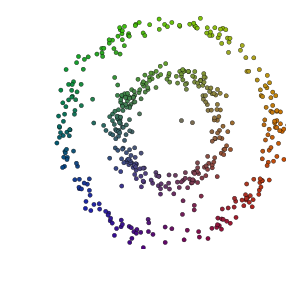}
	\caption{{\rmfamily\scshape Rad} on two circles.}
  \end{subfigure}
  \begin{subfigure}[t]{0.16\textwidth}
	\centering
    \includegraphics[width=.8\textwidth]{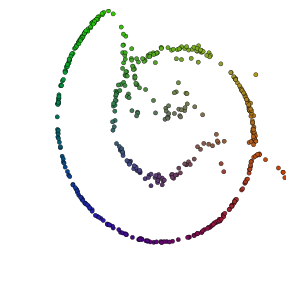}
	\caption{{\rmfamily\scshape Rad} on spiral.}
  \end{subfigure}
  \begin{subfigure}[t]{0.16\textwidth}
	\centering
    \includegraphics[width=.8\textwidth]{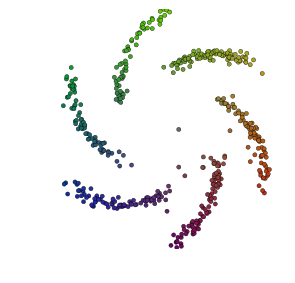}
	\caption{{\rmfamily\scshape Rad} on many moons.}
  \end{subfigure}
   \caption{Comparison of samples from trained
   {\rmfamily\scshape Real NVP} (top row) (a-f) and {\rmfamily\scshape Rad}
   (bottow row) (g-l) models. {\rmfamily\scshape Real NVP} fails in a low capacity
   setting by attributing probability mass over spaces where the data
   distribution has low density. Here, these spaces often connect
   data clusters, illustrating the challenges that come with modeling multimodal data as one continuous manifold.}
   \label{fig:toy_2d_samples}
\end{figure}

We plot the samples (Figure~\ref{fig:toy_2d_samples}) of the described
{\rmfamily\scshape Rad} and {\rmfamily\scshape Real NVP} models trained on
these problems. In the described low capacity regime,
{\rmfamily\scshape Real NVP} fails by attributing probability mass over spaces
where the data distribution has low density. This is consistent with the {\em mode covering}
behavior of maximum likelihood. However, the particular inductive bias of
{\rmfamily\scshape Real NVP} is to prefer modeling the data as one connected
manifold. This results in the unwanted probability mass being distributed along
the space between clusters.

\begin{figure}[t]
\captionsetup[subfigure]{justification=centering}
  \centering
  \begin{subfigure}[t]{0.47\textwidth}
	\centering
    \includegraphics[width=.5\textwidth]{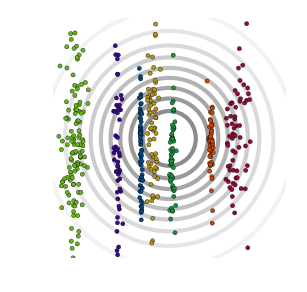}
	\caption{{\rmfamily\scshape Real NVP} Gaussianization.}
  \end{subfigure}
  \begin{subfigure}[t]{0.47\textwidth}
	\centering
    \includegraphics[width=.5\textwidth]{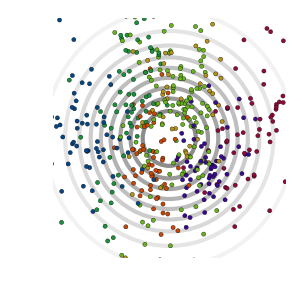}
	\caption{{\rmfamily\scshape Rad} Gaussianization.}
  \end{subfigure}
   \caption{Comparison of the Gaussianization process for
   {\rmfamily\scshape Rad} and {\rmfamily\scshape Real NVP} on the ring Gaussian
   mixture problem. Both plots show the image of data samples in the latent $\vz$ variables, 
   with level sets of the standard normal distribution plotted
   for reference. {\rmfamily\scshape Real NVP} leaves some area of this
   Gaussian unpopulated, an effect which is not visually apparent for
   {\rmfamily\scshape Rad}.}
   \label{fig:toy_2d_gaussianization}
\end{figure}

Flow-based models often follow the principle of
{\em Gaussianization}~\citep{chen2000gaussianization}, i.e. transforming the
data distribution into a Gaussian. The inversion of that process on a Gaussian
distribution would therefore approximate the data distribution. We plot in
Figure~\ref{fig:toy_2d_gaussianization} the inferred Gaussianized variables
$\vz^{(5)}$ for both models trained on the ring Gaussian mixture problem. The
Gaussianization from {\rmfamily\scshape Real NVP} leaves some area of the
standard Gaussian distribution unpopulated. These unattended areas correspond
to unwanted regions of probability mass in the input space.
{\rmfamily\scshape Rad} suffers significantly less from this problem.

An interesting feature is that {\rmfamily\scshape Rad} seems also to outperform
{\rmfamily\scshape Real NVP} on the spiral dataset. One hypothesis is that the
model successfully exploits some non-linear symmetries in this problem.

\subsubsection{Folding}
\begin{figure}[t]
\captionsetup[subfigure]{justification=centering}
  \centering
  \begin{subfigure}[t]{0.95\textwidth}
	\centering
    \includegraphics[width=\textwidth]{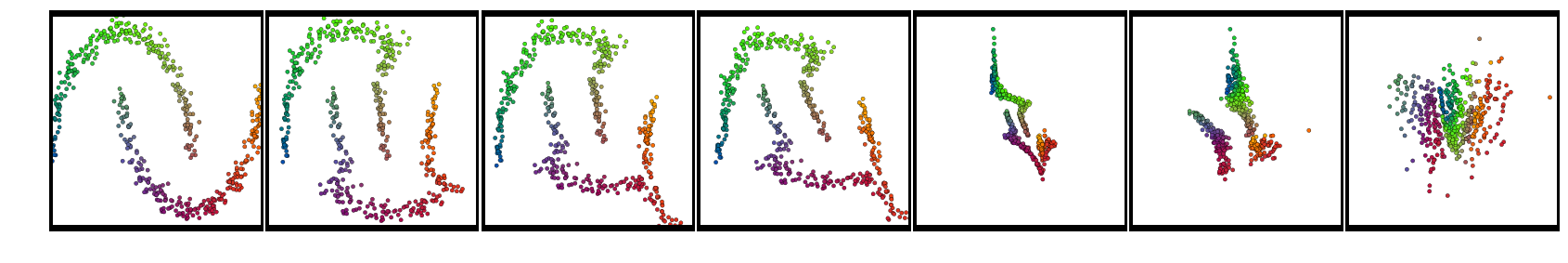}
	\caption{{\rmfamily\scshape Real NVP} inference.}
  \end{subfigure}
  \begin{subfigure}[t]{0.95\textwidth}
	\centering
    \includegraphics[width=\textwidth]{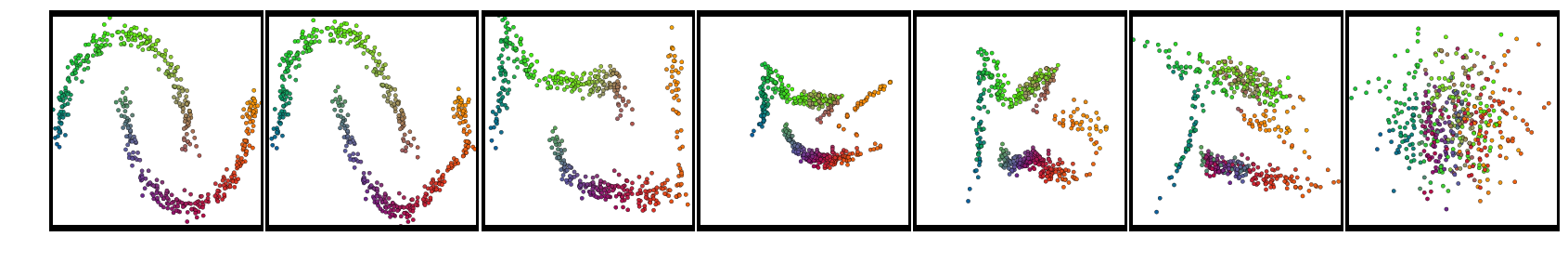}
	\caption{{\rmfamily\scshape Rad} inference.}
  \end{subfigure}
   \caption{Comparison of the inference process for {\rmfamily\scshape Rad}
   and {\rmfamily\scshape Real NVP} on the two moons problem. Each pane shows input samples embedded in different networks layers, 
   progressing from left to right from earlier to later network layers.
   The points are colored according to their original position in the input
   space. In
   {\rmfamily\scshape Rad} several points which were far apart in the input
   space become neighbors in $\vz^{(5)}$. This is not the case for
   {\rmfamily\scshape Real NVP}.}
   \label{fig:toy_2d_inference}
\end{figure}

We take a deeper look at the Gaussianization process involved in both models.
In Figure~\ref{fig:toy_2d_inference} we plot the inference process of $\vz^{(5)}$
from $\vx$ for both models trained on the two moons problem.
As a result of
a folding process similar to that in~\citet{montufar2014number}, several points
which were far apart in the input space become neighbors in $z^{(5)}$ in the
case of {\rmfamily\scshape Rad}.

\begin{figure}[t]
\captionsetup[subfigure]{justification=centering}
  \centering
  \begin{subfigure}[t]{0.32\textwidth}
    \centering
    \includegraphics[width=.6\textwidth]{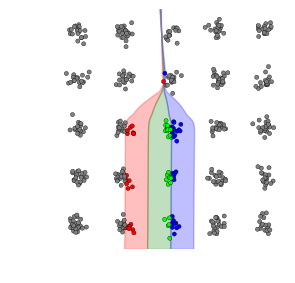}
    \caption{Input points of a {\rmfamily\scshape Rad} layer. The red, green,
      and blue colors corresponds to different labels of the partition subsets ($\abs{K}$ values),
      domains of $\sA_k$ for different $k$,
      where the function is non-invertible without knowing $k$ (see (b)).
      The black points are in the invertible area,
    where $k$ is not needed for the inversion.}
  \end{subfigure}
  \begin{subfigure}[t]{0.32\textwidth}
    \centering
    \includegraphics[width=.6\textwidth]{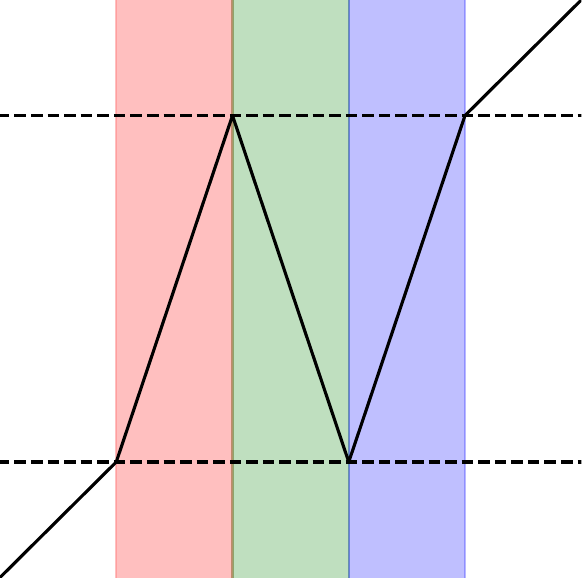}
    \caption{An example of piecewise linear function used in a
    {\rmfamily\scshape Rad} layer. The red, green, and blue colors corresponds
    to the different labels of the partition subsets in the non-invertible
    area. The dashed lines correspond to the non-invertible area in output
    space.}
  \end{subfigure}
  \begin{subfigure}[t]{0.32\textwidth}
    \centering
    \includegraphics[width=.6\textwidth]{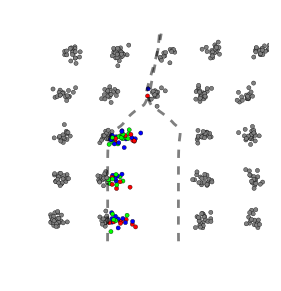}
    \caption{Output points of a {\rmfamily\scshape Rad} layer. The red, green,
    and blue colors corresponds to the different labels of the partition
    subsets in the non-invertible area of the input space, where points are
    folded on top of each other. The black points are
    in the invertible area, where $k$ is not needed for the inversion. The
    dashed lines correspond to the non-invertible area in output space.}
  \end{subfigure}
   \caption{Understanding the folding process, and understanding other visualizations of the folding process.
   }
   \label{fig:toy_2d_fold_viz}
\end{figure}

We further explore this folding process using the visualization described in
Figure~\ref{fig:toy_2d_fold_viz}. We verify that the non-linear folding process
induced by {\rmfamily\scshape Rad} plays at least two roles: bridging gaps in
the distribution of probability mass, and exploiting symmetries in the data.

\begin{figure}[t]
\vspace{-5pt}
\captionsetup[subfigure]{justification=centering}
  \centering
  \begin{subfigure}[t]{0.95\textwidth}
	\centering
    \includegraphics[width=.7\textwidth]{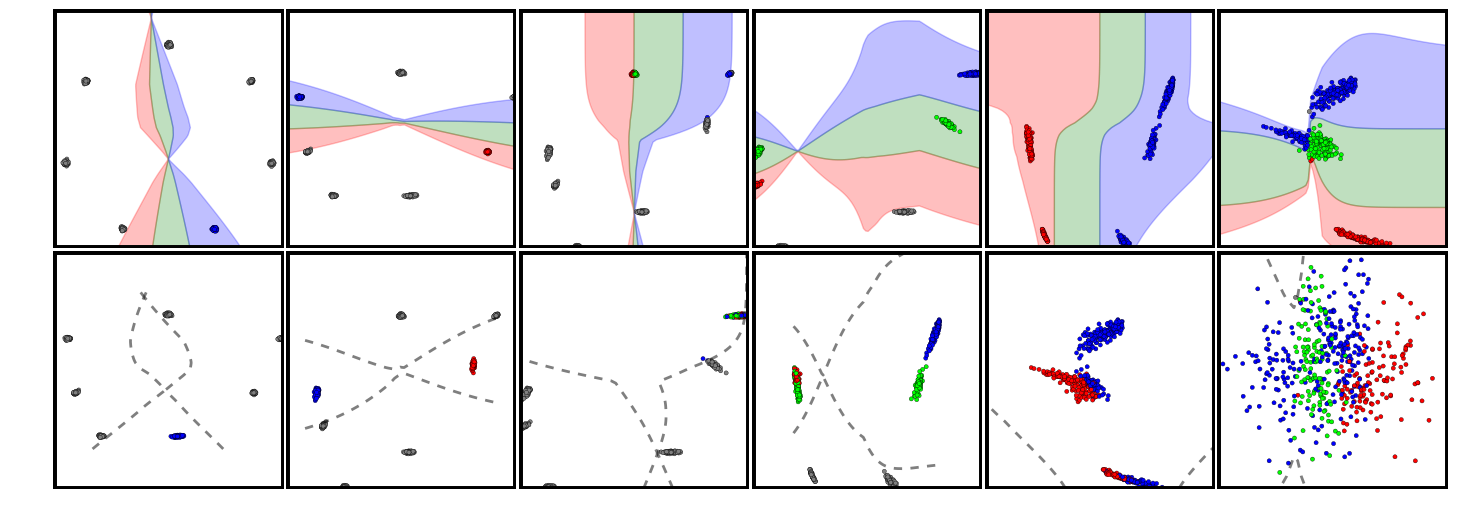}
	\caption{{\rmfamily\scshape Rad} folding strategy on the ring Gaussian
	mixture problem. The top rows correspond to each {\rmfamily\scshape Rad} layer's
	input points, and the bottom rows to its output points, as shown
	in~\ref{fig:toy_2d_fold_viz}. The labels tends to be well separated in
	output space as well.}
    \label{fig:ring_mog_rad}
  \end{subfigure}

  \begin{subfigure}[t]{0.95\textwidth}
	\centering
    \includegraphics[width=.7\textwidth]{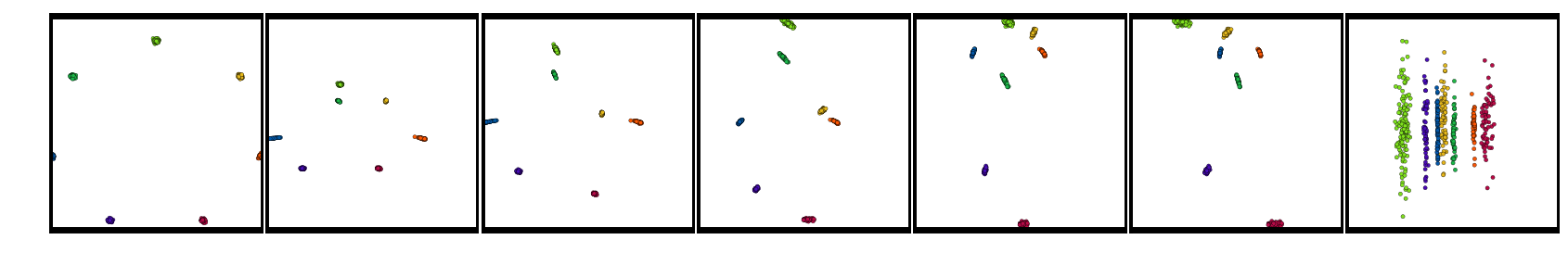}
	\caption{{\rmfamily\scshape Real NVP} inference strategy on the ring
	Gaussian mixture problem. The points are colored according to their original
	position in the input space.}
    \label{fig:ring_mog_rnvp}
  \end{subfigure}
   \caption{{\rmfamily\scshape Rad} and {\rmfamily\scshape Real NVP} inference
   processes on the ring Gaussian mixture problem. Each column correspond to a
   {\rmfamily\scshape Rad} or affine coupling layer. {\rmfamily\scshape Rad}
   effectively uses foldings in order to bridge the multiple modes of the
   distribution into a single mode, primarily in the last layers of the
   transformation, whereas {\rmfamily\scshape Real NVP} struggles to bring
   together these modes under the standard Gaussian distribution using continuous
   bijections.
   }
   \label{fig:ring_mog}
\vspace{-15pt}
\end{figure}

We observe that in the case of the ring Gaussian mixture
(Figure~\ref{fig:ring_mog_rad}), {\rmfamily\scshape Rad} effectively uses
foldings in order to bridge the different modes of the distribution into a
single mode, primarily in the last layers of the transformation. We contrast
this with {\rmfamily\scshape Real NVP} (Figure~\ref{fig:ring_mog_rnvp}) which
struggles to combine these modes under the standard Gaussian distribution
using bijections.

\begin{figure}[b]
\vspace{-15pt}
\captionsetup[subfigure]{justification=centering}
  \centering
  \begin{subfigure}[t]{0.95\textwidth}
	\centering
    \includegraphics[width=.7\textwidth]{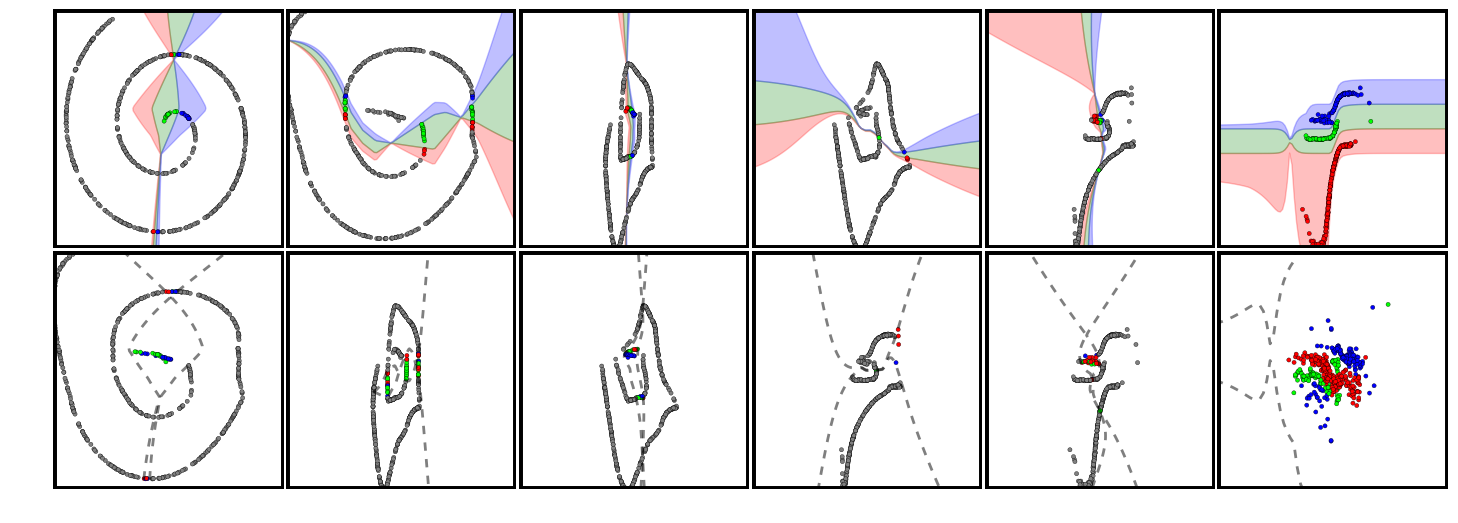}
	\caption{{\rmfamily\scshape Rad} folding strategy on the spiral problem.
	The top rows correspond to each {\rmfamily\scshape Rad} layer's input points,
	and the bottom rows to its output points, as shown in~\ref{fig:toy_2d_fold_viz}.}
    \label{fig:spiral_rad}
  \end{subfigure}

  \begin{subfigure}[t]{0.95\textwidth}
	\centering
    \includegraphics[width=.7\textwidth]{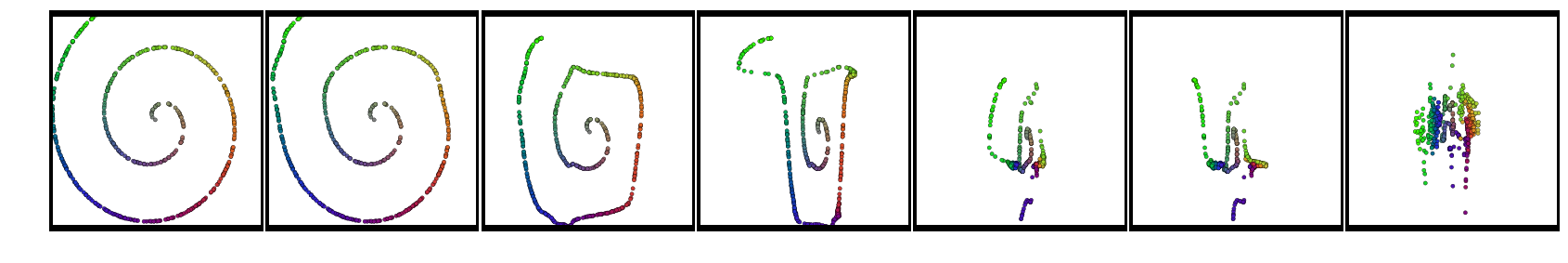}
	\caption{{\rmfamily\scshape Real NVP} inference strategy on the spiral
	problem. The points are colored according to their original position in the
	input space.}
    \label{fig:spiral_rnvp}
  \end{subfigure}
   \caption{{\rmfamily\scshape Rad} and {\rmfamily\scshape Real NVP} inference
   processes on the spiral problem. Each column correspond to a
   {\rmfamily\scshape Rad} or affine coupling layer. Instead of unrolling the
   manifold as {\rmfamily\scshape Real NVP} tries to, {\rmfamily\scshape Rad}
   uses a more successful strategy of decomposing the spiral into three
   different lines that it later bridges. While the function is surjective,
   {\rmfamily\scshape Rad} does not necessarily mix together points of
   different colors (original regions) together but can keep them separate:
   this is a more concrete illustration of the effect of $p_{K \mid Z}$
   shown in Figure~\ref{fig:gating}.
   }
   \label{fig:spiral}
\vspace{-5pt}
\end{figure}

In the spiral problem (Figure~\ref{fig:spiral}), {\rmfamily\scshape Rad}
decomposes the spiral into three different lines to bridge
(Figure~\ref{fig:spiral_rad}) instead of unrolling the manifold fully, which
{\rmfamily\scshape Real NVP} struggles to do (Figure~\ref{fig:spiral_rnvp}).

In both cases, the points remain generally well separated by labels, even after
being pushed through a {\rmfamily\scshape Rad} layer
(Figure~\ref{fig:ring_mog_rad} and~\ref{fig:spiral_rad}). This enables the
model to maximize the conditional log-probability $\log(p_{K \mid Z})$.

\section{Conclusion}
We introduced an approach to tractably evaluate and train deep mixture models
using surjective piecewise invertible maps as a folding mechanism.
This allows exact inference, exact generation, and exact evaluation of
log-likelihood, avoiding many issues in previous discrete variables models.
This method can easily be combined with other flow based architectural
components \cite[for a more comprehensive framework, read][]{nielsen2020survae},
allowing flow based models to better model datasets with discrete as well as
continuous structure.

\subsubsection*{Acknowledgements}
The authors would like to thank
Kyle Kastner,
Johanna Hansen,
Harm De Vries,
Ben Poole,
Prajit Ramachandran,
Dustin Tran,
Erin Grant,
David Grangier,
George J. Tucker,
Matt D. Hoffman,
Daniel Duckworth,
Anna Huang,
Arvind Neelakantan,
Dale Schuurmans,
Graham Taylor,
Bart van Merrienb\"oer,
Daniel Duckworth,
Vincent Dumoulin,
Didrik Nielsen,
Priyank Jaini,
Emiel Hoogeboom,
Kyunghyun Cho,
Marc G. Bellemare,
Ross Goroshin,
the {\em ICLR 2019 Deep Generative Models for Highly Structured Data} reviewers,
and the {\em AISTATS 2020} reviewers,
for valuable discussion and feedbacks on this paper. We also thank
Brandon Amos,
Mathieu Blondel,
and Misha Denil for writing advice on this section.

We would also like to thank
the Python community~\citep{van1995python,oliphant2007python} for developing
the tools that enabled this work, including
{NumPy}~\citep{oliphant2006guide,walt2011numpy},
{SciPy}~\citep{jones2001scipy},
{Matplotlib}~\citep{hunter2007matplotlib},
{Tensorflow}~\citep{abadi2016tensorflow},
and {JAX}~\citep{bradbury2018jax}.

\bibliography{local}
\bibliographystyle{iclr2019_conference}

\appendix

\section{Inference processes}
\begin{figure}[b]
\captionsetup[subfigure]{justification=centering}
  \centering
  \begin{subfigure}[t]{0.32\textwidth}
  \centering
    \includegraphics[width=.55\textwidth]{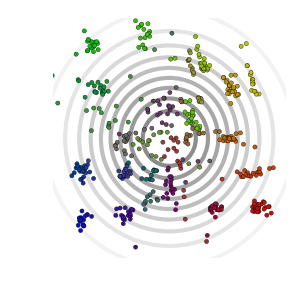}
  \caption{{\rmfamily\scshape Real NVP} on grid Gaussian mixture.}
  \end{subfigure}
  \begin{subfigure}[t]{0.32\textwidth}
  \centering
    \includegraphics[width=.55\textwidth]{toy_09_z_fin_color.png}
  \caption{{\rmfamily\scshape Real NVP} on ring Gaussian mixture.}
  \end{subfigure}
  \begin{subfigure}[t]{0.32\textwidth}
  \centering
    \includegraphics[width=.55\textwidth]{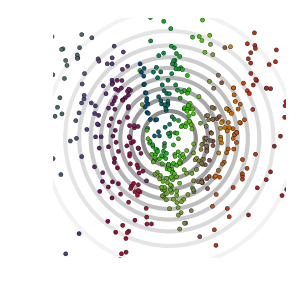}
  \caption{{\rmfamily\scshape Real NVP} on two moons.}
  \end{subfigure}
  \begin{subfigure}[t]{0.32\textwidth}
  \centering
    \includegraphics[width=.55\textwidth]{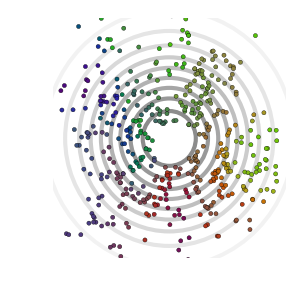}
  \caption{{\rmfamily\scshape Real NVP} on two circles.}
  \end{subfigure}
  \begin{subfigure}[t]{0.32\textwidth}
  \centering
    \includegraphics[width=.55\textwidth]{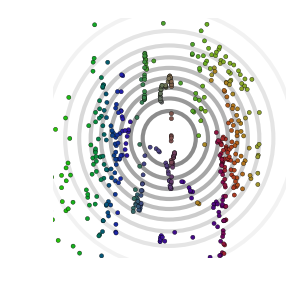}
  \caption{{\rmfamily\scshape Real NVP} on spiral.}
  \end{subfigure}
  \begin{subfigure}[t]{0.32\textwidth}
     \centering
    \includegraphics[width=.55\textwidth]{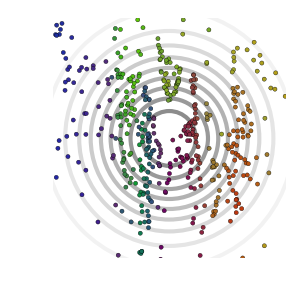}
    \caption{{\rmfamily\scshape Real NVP} on many moons.}
  \end{subfigure}

  \begin{subfigure}[t]{0.32\textwidth}
  \centering
    \includegraphics[width=.55\textwidth]{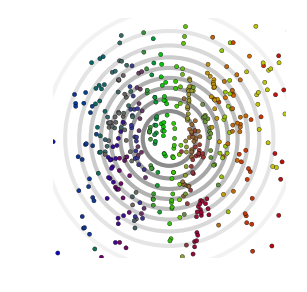}
  \caption{{\rmfamily\scshape Rad} on grid Gaussian mixture.}
  \end{subfigure}
  \begin{subfigure}[t]{0.32\textwidth}
  \centering
    \includegraphics[width=.55\textwidth]{toy_10_z_fin_color.png}
  \caption{{\rmfamily\scshape Rad} on ring Gaussian mixture.}
  \end{subfigure}
  \begin{subfigure}[t]{0.32\textwidth}
  \centering
    \includegraphics[width=.55\textwidth]{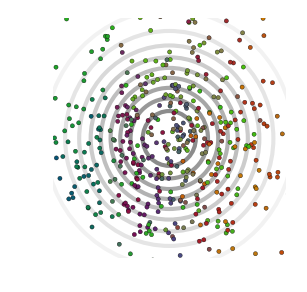}
  \caption{{\rmfamily\scshape Rad} on two moons.}
  \end{subfigure}
  \begin{subfigure}[t]{0.32\textwidth}
  \centering
    \includegraphics[width=.55\textwidth]{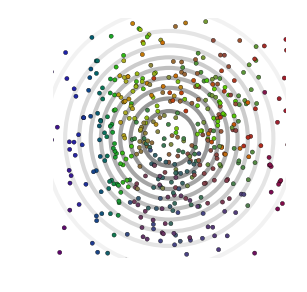}
  \caption{{\rmfamily\scshape Rad} on two circles.}
  \end{subfigure}
  \begin{subfigure}[t]{0.32\textwidth}
  \centering
    \includegraphics[width=.55\textwidth]{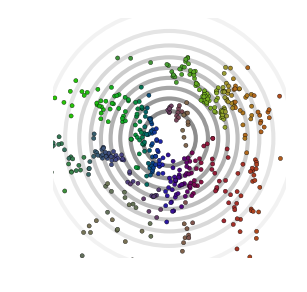}
  \caption{{\rmfamily\scshape Rad} on spiral.}
  \end{subfigure}
  \begin{subfigure}[t]{0.32\textwidth}
  \centering
    \includegraphics[width=.55\textwidth]{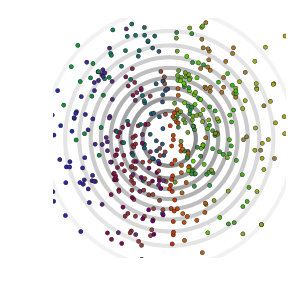}
  \caption{{\rmfamily\scshape Rad} on many moons.}
  \end{subfigure}
   \caption{Comparison of the Gaussianization from the trained
   {\rmfamily\scshape Real NVP} (top row) (a-f) and {\rmfamily\scshape Rad}
   (bottow row) (g-l). {\rmfamily\scshape Real NVP} fails in a low capacity
   setting by leaving unpopulated areas where the standard Gaussian attributes
   probability mass. Here, these spaces as often ones separating clusters,
   showing the failure in modeling the data as one manifold.}
   \label{fig:toy_2d_gaussianizations}
\end{figure}

\begin{figure}[t]
\captionsetup[subfigure]{justification=centering}
  \centering
  \begin{subfigure}[t]{0.95\textwidth}
	\centering
    \includegraphics[width=.75\textwidth]{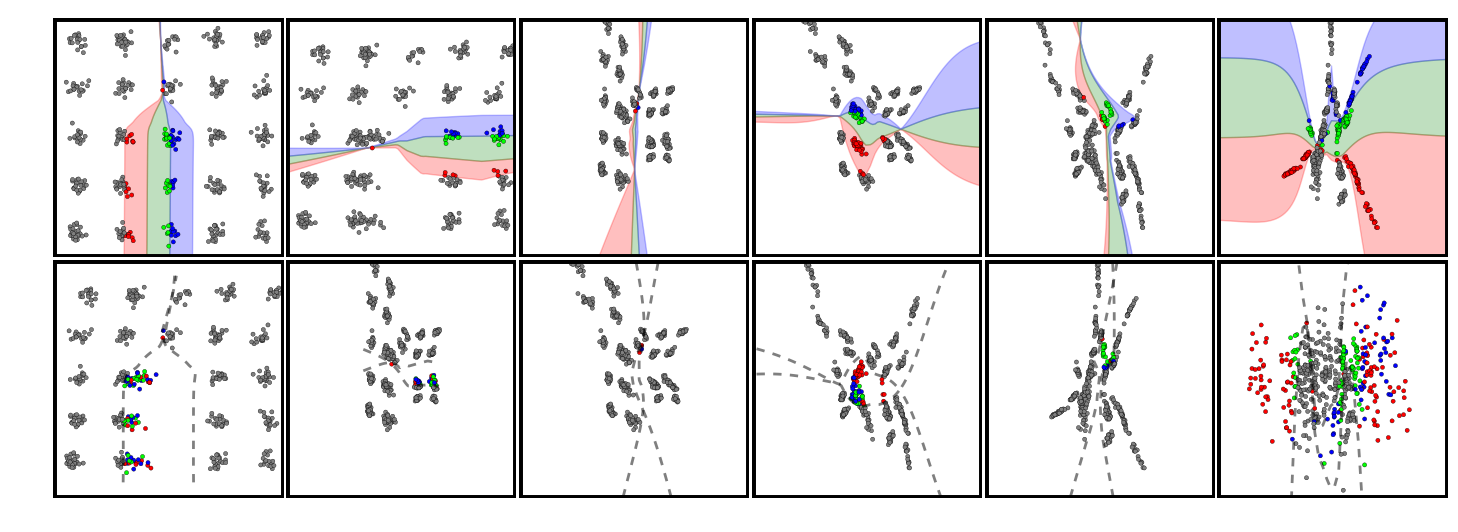}
	\caption{{\rmfamily\scshape Rad} folding strategy on the grid Gaussian
	mixture problem. The top rows correspond to a {\rmfamily\scshape Rad} layer
	input points, and the bottom rows to its output points, as shown in~\ref{fig:toy_2d_fold_viz}.}
    \label{fig:grid_gmm_rad}
  \end{subfigure}

  \begin{subfigure}[t]{0.95\textwidth}
	\centering
    \includegraphics[width=.75\textwidth]{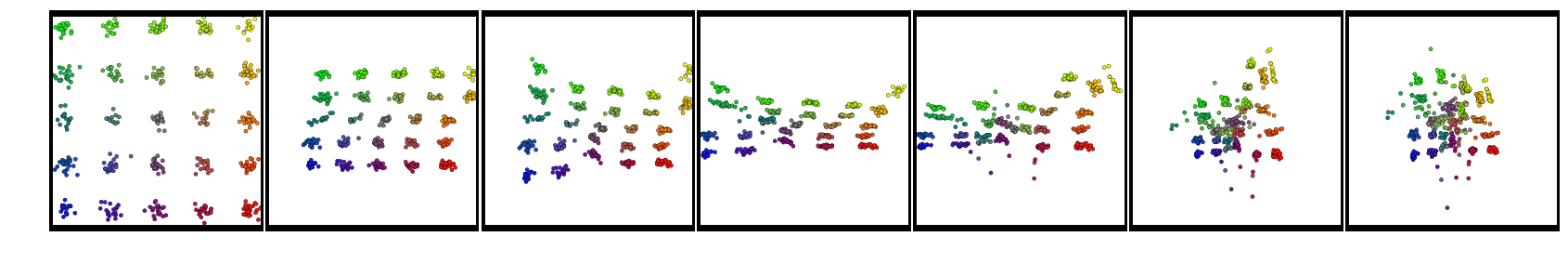}
	\caption{{\rmfamily\scshape Real NVP} inference strategy on the grid
	Gaussian mixture problem. The points are colored according to their original
	position in the input space.}
    \label{fig:grid_gmm_rnvp}
  \end{subfigure}
   \caption{{\rmfamily\scshape Rad} and {\rmfamily\scshape Real NVP} inference
   process on the grid Gaussian mixture problem. Each column correspond to a
   {\rmfamily\scshape Rad} or affine coupling layer.
   }
   \label{fig:grid_gmm}
\end{figure}

\begin{figure}[t]
\captionsetup[subfigure]{justification=centering}
  \centering
  \begin{subfigure}[t]{0.95\textwidth}
	\centering
    \includegraphics[width=.75\textwidth]{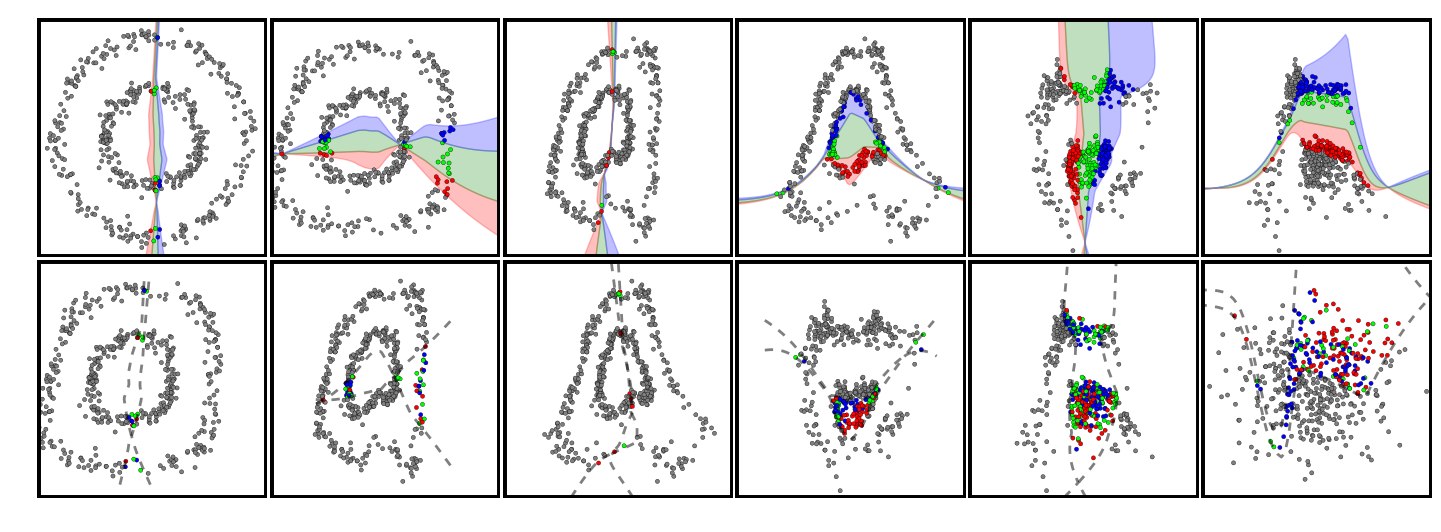}
	\caption{{\rmfamily\scshape Rad} folding strategy on the two circles
	problem. The top rows correspond to a {\rmfamily\scshape Rad} layer input
	points, and the bottom rows to its output points, as shown
	in~\ref{fig:toy_2d_fold_viz}.}
    \label{fig:circles_rad}
  \end{subfigure}

  \begin{subfigure}[t]{0.95\textwidth}
	\centering
    \includegraphics[width=.75\textwidth]{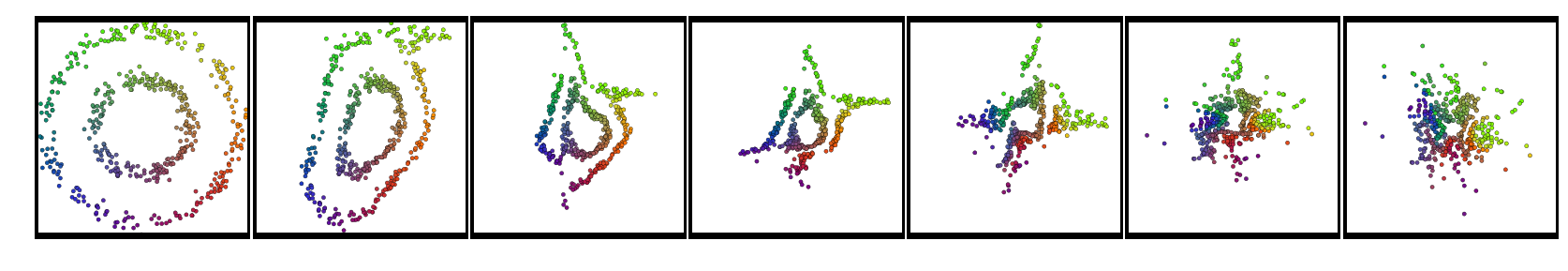}
	\caption{{\rmfamily\scshape Real NVP} inference strategy on the two circles
	problem. The points are colored according to their original position in the
	input space.}
    \label{fig:circles_rnvp}
  \end{subfigure}
   \caption{{\rmfamily\scshape Rad} and {\rmfamily\scshape Real NVP} inference
   process on the two circles problem. Each column correspond to a
   {\rmfamily\scshape Rad} or affine coupling layer.
   }
   \label{fig:circles}
\end{figure}

\begin{figure}[t]
\captionsetup[subfigure]{justification=centering}
  \centering
  \begin{subfigure}[t]{0.95\textwidth}
	\centering
    \includegraphics[width=.75\textwidth]{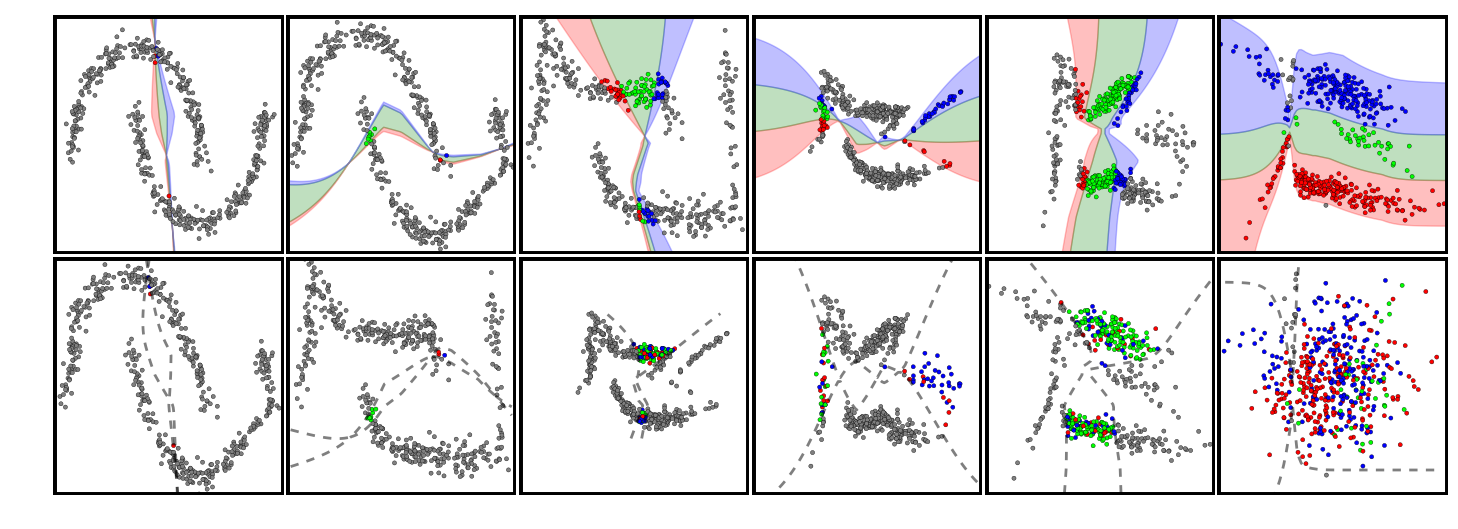}
	\caption{{\rmfamily\scshape Rad} folding strategy on the two moons problem.
	The top rows correspond to a {\rmfamily\scshape Rad} layer input points,
	and the bottom rows to its output points, as shown
	in~\ref{fig:toy_2d_fold_viz}.}
    \label{fig:2_moons_rad}
  \end{subfigure}

  \begin{subfigure}[t]{0.95\textwidth}
	\centering
    \includegraphics[width=.75\textwidth]{toy_01_zs.png}
	\caption{{\rmfamily\scshape Real NVP} inference strategy on the two moons
	problem. The points are colored according to their original position in the
	input space.}
    \label{fig:2_moons_rnvp}
  \end{subfigure}
   \caption{{\rmfamily\scshape Rad} and {\rmfamily\scshape Real NVP} inference
   process on the two moons problem. Each column correspond to a
   {\rmfamily\scshape Rad} or affine coupling layer.
   }
   \label{fig:2_moons}
\end{figure}

\begin{figure}[t]
\captionsetup[subfigure]{justification=centering}
  \centering
  \begin{subfigure}[t]{0.95\textwidth}
	\centering
    \includegraphics[width=.75\textwidth]{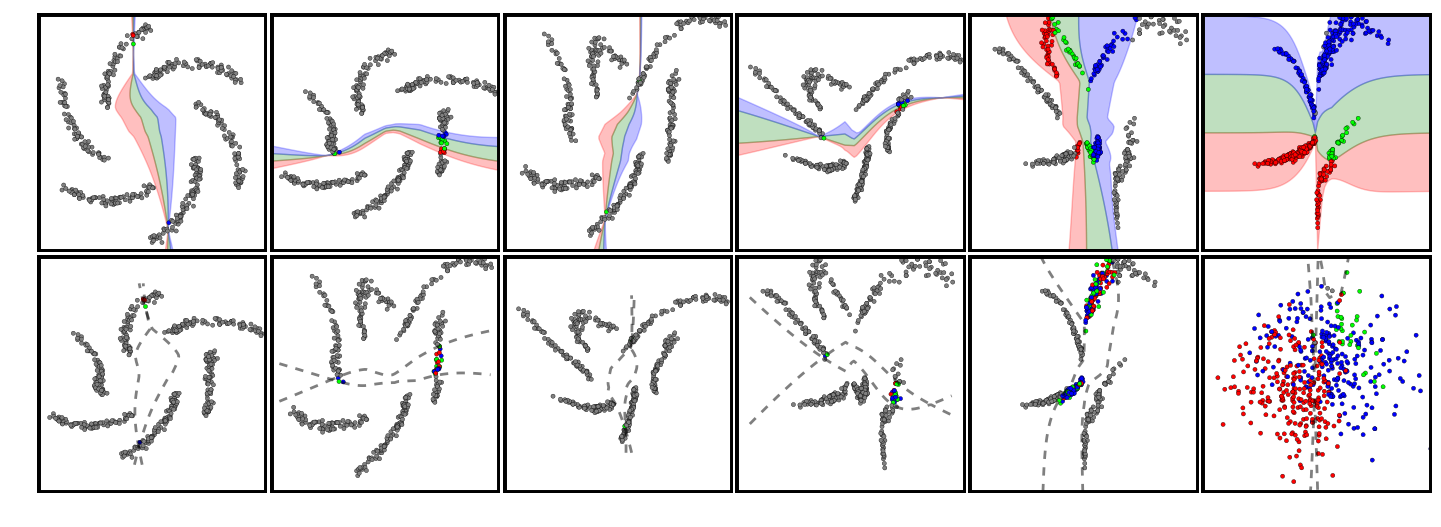}
	\caption{{\rmfamily\scshape Rad} folding strategy on the many moons
	problem. The top rows correspond to a {\rmfamily\scshape Rad} layer input
	points, and the bottom rows to its output points, as shown
	in~\ref{fig:toy_2d_fold_viz}.}
    \label{fig:many_moons_rad}
  \end{subfigure}

  \begin{subfigure}[t]{0.95\textwidth}
	\centering
    \includegraphics[width=.75\textwidth]{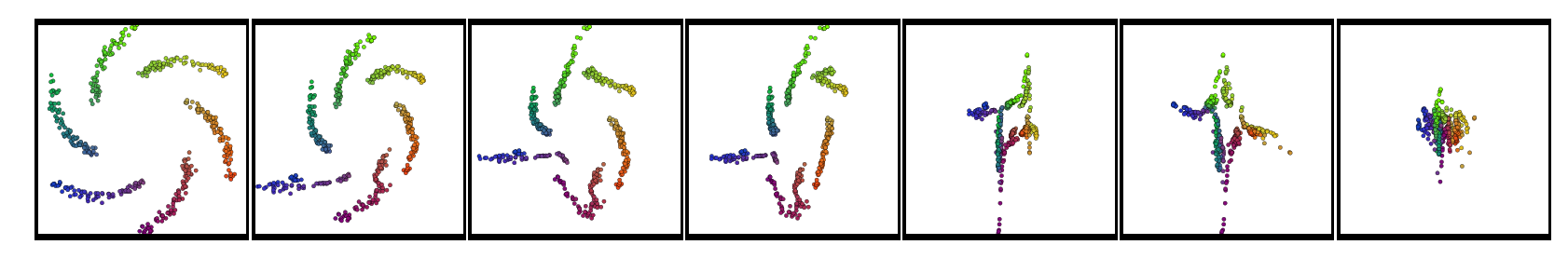}
	\caption{{\rmfamily\scshape Real NVP} inference strategy on the many moons
	problem. The points are colored according to their original position in the
	input space.}
    \label{fig:many_moons_rnvp}
  \end{subfigure}
   \caption{{\rmfamily\scshape Rad} and {\rmfamily\scshape Real NVP} inference
   process on the many moons problem. Each column correspond to a
   {\rmfamily\scshape Rad} or affine coupling layer.
   }
   \label{fig:many_moons}
\end{figure}

We plot the remaining inference processes of {\rmfamily\scshape Rad} and
{\rmfamily\scshape Real NVP} on the remaining problems not plotted previously:
{\em grid Gaussian mixture} (Figure~\ref{fig:grid_gmm}),
{\em two circles} (Figure~\ref{fig:circles}),
{\em two moons} (Figure~\ref{fig:2_moons}),
and {\em many moons} (Figure~\ref{fig:many_moons}). We also compare the final
results of the Gaussianization processes on both models on the different toy
problems in Figure~\ref{fig:toy_2d_gaussianizations}.

\end{document}

%% file: math_commands.tex
\usepackage{amsmath,amsfonts,bm}

\def\eqref#1{equation~\ref{#1}}

\def\1{\bm{1}}

\def\vx{{\bm{x}}}

\def\vz{{\bm{z}}}

\DeclareMathAlphabet{\mathsfit}{\encodingdefault}{\sfdefault}{m}{sl}
\SetMathAlphabet{\mathsfit}{bold}{\encodingdefault}{\sfdefault}{bx}{n}

\def\gU{{\mathcal{U}}}

\def\sA{{\mathbb{A}}}

\def\sR{{\mathbb{R}}}

\newcommand{\R}{\mathbb{R}}

